\PassOptionsToPackage{capitalize,noabbrev}{cleveref}

\documentclass[]{preprint}

\usepackage[toc,page,header]{appendix}

\usepackage[T1]{fontenc}
\usepackage{listings}
\usepackage{tcolorbox}
\tcbuselibrary{listings,breakable}
\usepackage{xcolor}
\usepackage{upquote}
\usepackage{circledsteps}

\usepackage{microtype}
\usepackage{graphicx}
\usepackage{subcaption}
\usepackage{booktabs} 

\usepackage{hyperref}

\usepackage{amsmath}
\usepackage{amssymb}
\usepackage{mathtools}
\usepackage{amsthm}

\usepackage{stmaryrd}

\theoremstyle{plain}

\theoremstyle{definition}

\theoremstyle{remark}

\usepackage{hyperref}
\usepackage{url}
\usepackage{xcolor}
\usepackage{pifont}

\usepackage[utf8]{inputenc}
\usepackage{csquotes} 
\usepackage{enumitem}
\usepackage{multirow}
\usepackage{array}
\usepackage{colortbl}

\usepackage{arydshln} 

\usepackage{wrapfig}
\usepackage{subcaption}  
\usepackage{dblfloatfix}

\usepackage[most]{tcolorbox}
\usepackage{xcolor}
\usepackage{graphicx}

\definecolor{cvprblue}{RGB}{0,102,204} 

\usepackage{amsmath,bm}

\usepackage{color}
\usepackage{tikz}
\usetikzlibrary{shapes, arrows.meta, positioning, calc, fit, backgrounds}

\usepackage{etoc}

\definecolor{sybcblue}{HTML}{004488}  
\definecolor{brickred}{HTML}{BB5566}  
\definecolor{softgray}{HTML}{666666}  

\definecolor{graybg}{gray}{0.9}

\newcommand{\second}[1]{\underline{#1}}      

\usepackage{hyperref}
\hypersetup{
    colorlinks=true,
    linkcolor=black,
    filecolor=magenta,      
    urlcolor=seedblue, 
    pdftitle={Modality Gap},
}

\usepackage{tikz}

\usepackage{cleveref}

\definecolor{softred}{RGB}{255,230,230}
\definecolor{softblue}{RGB}{230,240,255}
\definecolor{slowcolor}{RGB}{0,100,200} 

\usepackage{amsmath}   
\usepackage{amssymb}   
\usepackage{amsthm}    

\usepackage{tcolorbox}

\usepackage{enumitem}

\usepackage{amsmath,amssymb,amsthm,mathtools}

\usepackage[utf8]{inputenc} 
\usepackage[T1]{fontenc}    
\usepackage{hyperref}       
\usepackage{url}            
\usepackage{booktabs}       
\usepackage{amsfonts}       
\usepackage{nicefrac}       
\usepackage{microtype}      
\usepackage{xcolor}         

\usepackage[utf8]{inputenc} 
\usepackage[T1]{fontenc}    
\usepackage{hyperref}       
\usepackage{url}            
\usepackage{booktabs}       
\usepackage{amsfonts}       
\usepackage{nicefrac}       
\usepackage{microtype}      
\usepackage{xcolor}         

\usepackage{tikz}
\usetikzlibrary{arrows.meta, calc, shapes.geometric, decorations.pathreplacing, positioning}

\usepackage[utf8]{inputenc} 
\usepackage[T1]{fontenc}    
\usepackage{url}            
\usepackage{booktabs}       
\usepackage{amsfonts}       
\usepackage{nicefrac}       
\usepackage{microtype}      
\usepackage{xcolor}         

\usepackage{amsmath}
\usepackage{amssymb}

\usepackage[utf8]{inputenc} 
\usepackage[T1]{fontenc}    
\usepackage{url}            
\usepackage{booktabs}       
\usepackage{amsfonts}       
\usepackage{nicefrac}       
\usepackage{microtype}      

\usepackage{wrapfig}

\usepackage{pifont}

\usepackage[table]{xcolor}
\usepackage{booktabs}
\usepackage{multirow}
\usepackage{arydshln}

\usepackage{booktabs}
\usepackage{tabularx}

\usepackage{tikz}
\usetikzlibrary{spy, arrows.meta, calc, decorations.pathreplacing, bending, shadows} 

\usepackage{amsmath,amssymb}
\usepackage{tikz}
\usetikzlibrary{
  arrows.meta,
  positioning,
  calc,
  fit,
  backgrounds,
  decorations.pathreplacing,
  shapes.geometric,
  shapes.misc,
  matrix
}

\usepackage{tikz}

\usepackage{xcolor}

\usetikzlibrary{
    arrows.meta,      
    calc,             
    shapes.geometric, 
    fadings,          
    bending           
}

\usepackage{tikz}
\usetikzlibrary{arrows.meta, calc, decorations.pathreplacing, positioning}
\usepackage{amsmath, amssymb}

\usepackage{tikz}
\usetikzlibrary{arrows.meta, calc, angles, quotes}

\usepackage{booktabs}
\usepackage{multirow}
\usepackage{graphicx}
\usepackage[table]{xcolor}

\definecolor{graybg}{gray}{0.92}

\definecolor{promptblue}{HTML}{4F78B6}

\tcbset{
  promptbox/.style={
    enhanced,
    breakable,
    colframe=promptblue,
    colback=promptblue!3,
    opacityback=0.95,
    boxrule=0.7pt,
    arc=4pt,
    left=6pt,
    right=6pt,
    top=5pt,
    bottom=5pt,
    fonttitle=\bfseries,
    coltitle=white,
    colbacktitle=promptblue,
    boxed title style={
      arc=3pt,
      boxrule=0pt,
      left=5pt,
      right=5pt,
      top=2pt,
      bottom=2pt
    },
    attach boxed title to top left={
      xshift=6pt,
      yshift=-2pt
    }
  }
}

\usepackage{tikz}
\usetikzlibrary{arrows.meta, calc}
\usepackage{amsmath, amssymb}

\definecolor{src}{HTML}{1B4965}
\definecolor{tgt}{HTML}{C44536}
\definecolor{gold}{HTML}{B8860B}
\definecolor{ell}{HTML}{C5D8E8}
\definecolor{gridg}{HTML}{DCDCDC}
\definecolor{note}{HTML}{6E7F8E}

\definecolor{CaseBlue}{HTML}{2471A3}

\renewcommand{\abstractboxlogo}{%
  \includegraphics[width=2.8cm]{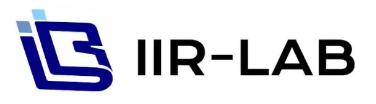}%
}

\title{ToolArtist: Tool-Using Unified Multimodal Models for Agentic Image Generation}
\newcommand{\ourmodel}{ToolArtist}

\newcommand{\corrauth}{\ding{41}}
\newcommand{\equalcontrib}{*}

\makeatletter
\renewcommand\author[2][]{\addtolist[#1]{#2}{\authorlist}{\authorformat}{, }}
\makeatother

\author[1]{Jiahao Zhao \equalcontrib}
\author[2,3]{Xiaomin Yu \equalcontrib}
\author[1]{Zhongxiang Sun}
\author[4]{Fengwei Teng}

\makeatletter
\renewcommand\author[2][]{\addtolist[#1]{#2}{\authorlist}{\authorformat}{\\[4pt]}}
\makeatother

\author[2]{Chengwei Qin}

\makeatletter
\renewcommand\author[2][]{\addtolist[#1]{#2}{\authorlist}{\authorformat}{, }}
\makeatother

\author[3]{Xiaobin Hu \corrauth}
\author[1]{Jun Xu \corrauth}
\author[3]{Shuicheng Yan}

\affiliation[1]{RUC}
\affiliation[2]{HKUST(GZ)}
\affiliation[3]{NUS}
\affiliation[4]{UCD}

\newcommand{\emailaddr}[1]{%
  \href{mailto:#1}{\nolinkurl{#1}}%
}

\providecommand{\firstpagefootnotes}{}
\renewcommand{\firstpagefootnotes}{%
  \parbox{0.96\textwidth}{%
    {\color{seedblue}\hrule height 0.2pt\relax}
    \vspace{6pt}
    \raggedright\normalsize
    $^{*}$ Equal contribution. \\
    \corrauth{} Corresponding authors \\
    Emails: \emailaddr{zhaojiahao2202@ruc.edu.cn}, \emailaddr{yuxm02@gmail.com}.
  }%
}

\abstract{
Text-to-image (T2I) models can produce visually compelling images, yet they remain limited on open-world tasks that require complex semantic understanding, multi-step reasoning, and the integration of external world knowledge. Existing efforts introduce agent capabilities into image generation, but they either prescribe a fixed workflow or place only a subset of the open-world image generation process under agent control. Consequently, reasoning, tool invocation, and image generation are not coordinated by a single policy. We propose \ourmodel, a fully \textbf{agentic image generation} model obtained by post-training a Unified Multimodal Model (UMM). \ourmodel{} dynamically orchestrates reasoning, external tool use, and native image generation within one unified policy. During Supervised Fine-Tuning (SFT), we equip a teacher agent with search tools alongside an image-generation tool. We then convert the collected trajectories into a UMM compatible format, where the image-generation tool is concealed while the resulting generated images are retained. During Reinforcement Learning (RL), we develop an agentic RL infrastructure for UMMs and introduce Reason-Act-Draw GRPO (RAD-GRPO), which uses complementary intent and quality rewards to jointly optimize the model. Experiments show that placing the entire open-world image-generation process under an agent policy consistently outperforms approaches with fixed pipelines or only partially agent-controlled components. We release the training data and the complete post-training infrastructure.
}

\date{\today}
\correspondence{Jun Xu, Xiaobin Hu}

\checkdata[Huggingface]{\textcolor{magenta}{\url{https://huggingface.co/datasets/bubble65/EMU-Agentic-PostTrain-Data}}}
\checkdata[Github]{\textcolor{magenta}{\url{https://github.com/bubble65/EMU-Agentic-PostTrain}}}

\begin{document}
\maketitle

\newpage


\section{Introduction}
\begin{figure}[h]
    \centering
    \includegraphics[width=\linewidth]{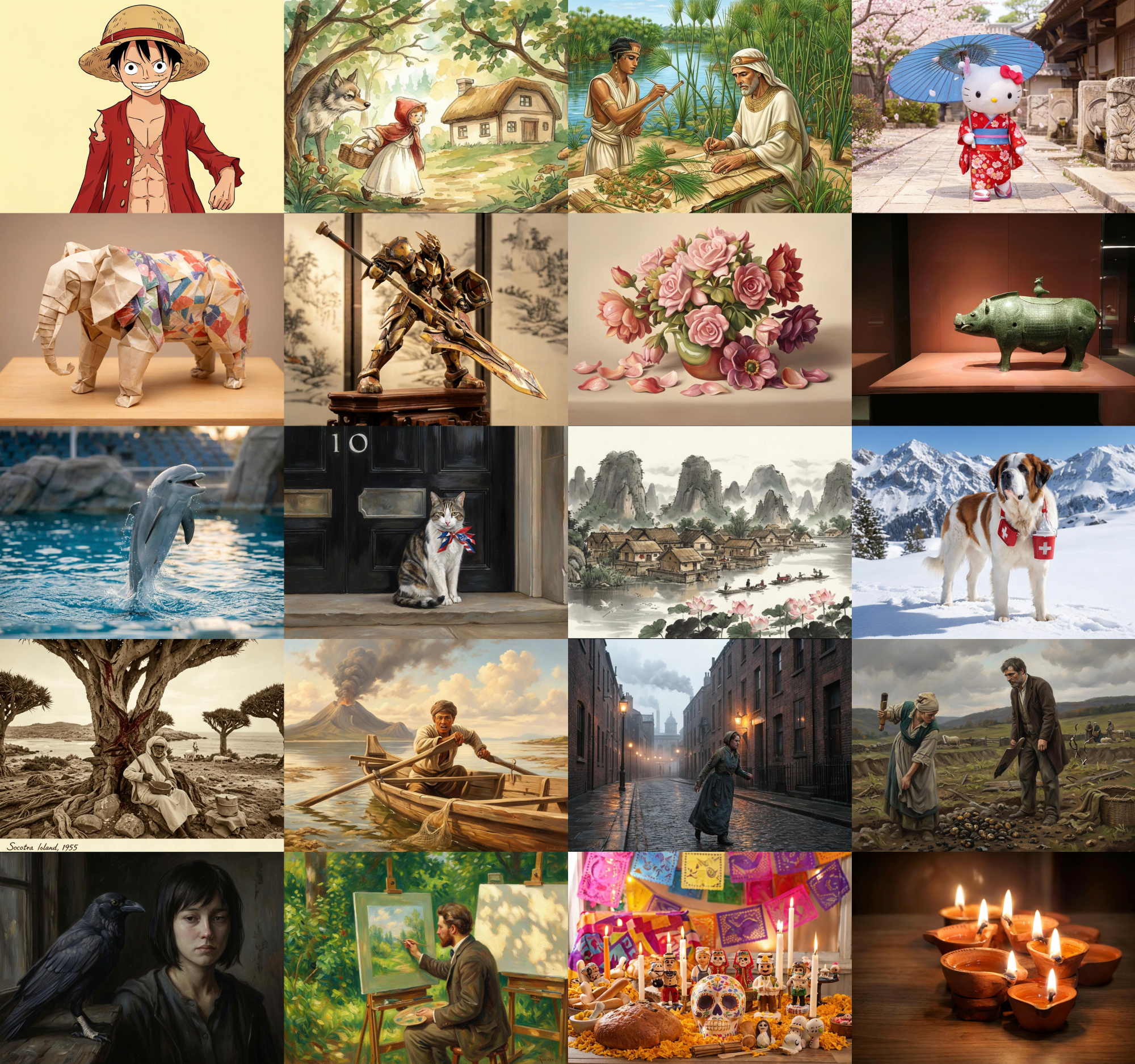}
    \caption{Demonstrations of open-world image generation, covering cultural knowledge, specific IPs, historical knowledge, geographic knowledge and more. All showcased images are generated by \textbf{\ourmodel}.}
    \label{fig:open_world_examples}
\end{figure}

Text-to-image (T2I) models~\cite{greenberg2025flux, podell2023sdxl, chen2023pixart, li2024playgroundv25, meituan2025longcatimage} can generate images with remarkable structure, detail and aesthetic quality. Nevertheless, these advances fail to enable reliable open-world image generation. For requests requiring complex semantic understanding, multi-hop reasoning, and knowledge of long-tail concepts, strict factual constraints or highly time-sensitive information, models often output visually plausible yet factually erroneous images. Recent benchmarks~\cite{han2025wordspixel, zhang2025apb, Huang2025t2ifactualbench, huang2025kitten, niu2026wise, fu2024commonsenset2i, zhang2025worldgenbench}, including WISE~\cite{niu2026wise} and WorldGenBench~\cite{zhang2025worldgenbench}, reveal this capability gap. Critical information is frequently omitted from user prompts and cannot be reliably stored within the model’s static parameters.

To overcome this challenge, contemporary works endow image‑generation systems with agentic capabilities to gather supplementary information via real‑world interaction, either intrinsically or with help from auxiliary agents.~\cite{wang2026searchgen, feng2026gensearcher, chen2026unifyagent, wu2023visualchatgpt, yang2024idea2img, shalevarkushin2026imagerag, zhang2026qwenimageagent, ye2026genclaw, jiang2026genagent, chen2026genevolve, chen2022reimagen} They broadly follow two paradigms. The first organizes prompt understanding, search, evidence aggregation, and image synthesis into a \textbf{predefined fixed} pipeline~\cite{chen2026unifyagent}. The second depends on a search agent~\cite{zhao2026dllmsearcher, song2025r1searcher} to gather evidence and optimize the user instruction into a grounded generation prompt, which is then passed to an \textbf{external} image generator~\cite{feng2026gensearcher, chen2026genevolve, jiang2026genagent}. These approaches demonstrate that introducing agentic capabilities can substantially improve open-world image generation quality.

However, they incorporate only part of the agentic capability required by open-world image generation. In the prompt optimization paradigm, the learned agent terminates at generation and delegates synthesis to a separate generator. In the pipeline paradigm, the order and role of search and generation are prescribed in advance. Under both paradigms, image generation is not an action freely chosen and initiated by the model. Consequently, the model does not learn the complete decision process from identifying a knowledge gap, through selecting and using external tools, to producing the final visual output.

\emph{\textbf{Fully agentic image generation requires tool use and image generation to be autonomous actions of the same policy.}}

We propose \textbf{\ourmodel}, a fully agentic image-generation model obtained by post-training a Unified Multimodal Model (UMM)~\cite{chen2025janus,wang2024emu3,cui2025emu35,tencent2026hunyuanimage30,chen2025blip3o,deng2025bagel}. Given a user request, ToolArtist reasons about missing information, decides whether and how to invoke external tools, incorporates the returned textual and visual evidence into its evolving multimodal context, and natively generates and revises the requested image. Image generation is therefore a part of the agent policy.

\begin{figure}[t]
    \centering
    \includegraphics[width=\linewidth]{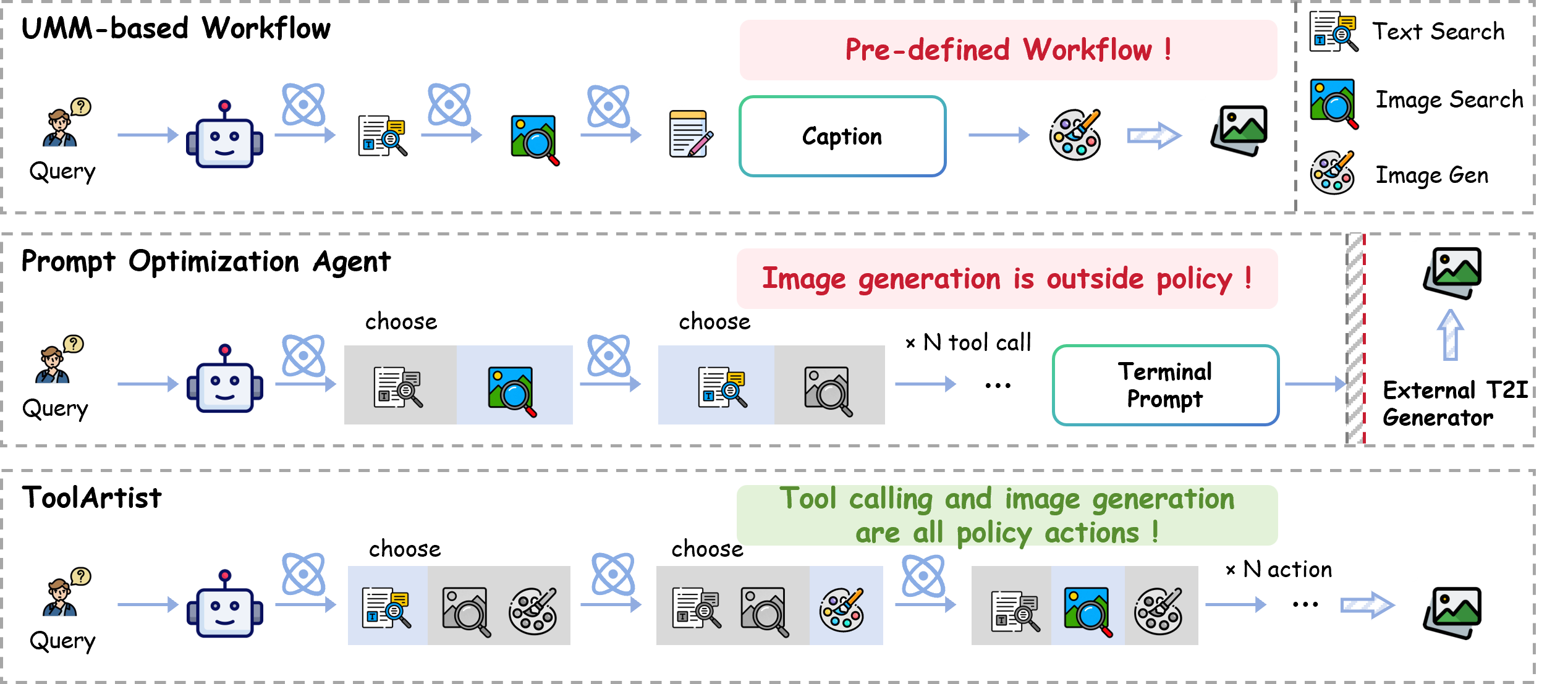}
    \caption{
\textbf {Comparison of Agentic Image Generation Paradigms.}
\textbf{Top:} The UMM-based Workflow paradigm introduces tool calling to the UMM, and the model also completes the final image generation. However, the workflow is predefined in advance.
\textbf{Middle:} The Prompt Optimization Agent employs a search agent to rewrite user instructions, and finally feeds the optimized prompt to an external generator.
\textbf{Bottom:} Our ToolArtist. Both tool calling and image generation are fully determined by the model itself.
}
    \label{fig:policy_workflow}
\end{figure}

We develop a post-training strategy to acquire this capability. During SFT, we equip a teacher agent with text-search and image-search tools together with an image-generation tool, allowing it to autonomously construct complete trajectories. We then convert the collected trajectories into a UMM-compatible format: the explicit image-generation tool is concealed, while each generated image is retained and represented as native visual-caption and visual-token spans. This conversion transfers the teacher's tool-using behavior into a form in which generation is performed by the UMM itself. Using only 7,132 high-quality trajectories, SFT establishes the model's basic ability to reason, invoke search tools, use retrieved evidence, and generate images within one policy.

During RL, we construct a complete UMM agentic RL infrastructure and introduce RAD-GRPO to optimize full agentic image-generation trajectories online. RAD-GRPO incorporates two complementary signals. \ding{182} Intent Reward evaluates whether the final generation description accurately and sufficiently translates the user request and acquired evidence into an executable visual intent. \ding{183} Quality Reward evaluates whether the generated image is faithful to that intent and the original request while maintaining high visual quality. Their combination propagates outcome-level feedback across reasoning, tool use, evidence utilization, and native image generation, further unlocking the agentic capabilities initialized by SFT.

We evaluate ToolArtist on WISE and WorldGenBench-Humanities. The results show that our fully agentic formulation outperforms open-source UMMs and image-generation methods that incorporate only partial agentic capabilities. We release the training data and the complete SFT and RL infrastructure to facilitate future research on agentic image generation.

Our contributions are as follows:

\textbf{1. Fully Agentic Image Generation Paradigm} \ourmodel, a fully agentic image-generation model that unifies autonomous reasoning, external tool utilization, and native visual generation under a single UMM policy.

\textbf{2. Post-training Strategy for UMMs} We design a post-training strategy to endow UMMs with agentic image generation capabilities, achieving better performance than approaches with only partial agentic functionalities.

\textbf{3. High-quality Open-source Data and Complete Infrastructure} We construct a dataset of 7k high-quality SFT trajectories and build a full training infrastructure supporting agentic SFT and RL for UMMs. All resources are released publicly.


\section{Preliminary}

\subsection{Open-World Image Generation}

Conventional T2I generation is typically formulated as a closed-world conditional generation problem. Given a user instruction $q$, a generator $G$ directly samples an image $\mathcal{I}$ from a prompt-conditioned distribution. This formulation assumes that all information required for generation is either explicitly provided in $q$ or implicitly stored in the parameters of $G$. However, this assumption breaks down when the target image depends on external information that must be acquired from the world. We term this setting \textbf{open-world image generation}. Let $\mathcal{W}$ denote the world-knowledge space and $\mathcal{Z}\subseteq\mathcal{W}$ the task-relevant textual and visual evidence. Rather than being available in advance, $\mathcal{Z}$ must be actively discovered from the environment:
\begin{equation*}
\mathcal{I}\sim G(\cdot\mid q,\mathcal{Z})
\qquad
\mathcal{Z}\subseteq\mathcal{W}.
\end{equation*}

\subsection{Unified Multimodal Model: Emu3.5}
UMMs represent both language and images as tokens and model them within a single multimodal sequence. We build ToolArtist on Emu3.5~\cite{cui2025emu35}, a native autoregressive UMM trained by unified next-token prediction over interleaved vision--language data. Let $\mathcal{V}_{\mathrm{text}}$ and $\mathcal{V}_{\mathrm{visual}}$ denote its textual and visual token vocabularies. An interleaved sequence $\mathbf{x}=(x_1,\ldots,x_L)$ contains tokens from $\mathcal{V}_{\mathrm{text}}\cup\mathcal{V}_{\mathrm{visual}}$ and is modeled as
\begin{equation*}
p_\theta(\mathbf{x})
=
\prod_{j=1}^{L}p_\theta(x_j\mid x_{<j}).
\end{equation*}
Under this formulation, user instructions, reasoning traces, and tool calls are represented as textual tokens, while input images, images returned by tools, and model-generated images are represented as visual-token spans. Emu3.5 can therefore process multimodal observations and natively produce images within the same autoregressive context, making it suitable as a unified policy over reasoning, acting, and drawing.

\subsection{Agentic Image Generation}

Following the ReAct~\cite{yao2023react} paradigm, we formulate agentic image generation as an iterative interaction between a UMM and the open-world environment. We instantiate \ourmodel{} as
\begin{equation*}
A_\theta=
(\pi_\theta,\mathcal{T},\mathcal{W}),
\end{equation*}
where $\pi_\theta$ is the UMM, $\mathcal{T}=\{\texttt{TextSearch},\texttt{ImageSearch}\}$ is the set of tools, and $\mathcal{W}$ is the open-world environment. Upon receiving $q$, the policy performs a variable number of interaction rounds. At round $t$, it conditions on the complete multimodal history $\mathcal{H}_{t-1}$, produces a reasoning span $r_t$, and then selects an action $a_t$: either calling a tool from $\mathcal{A}_{\mathrm{tools}}$ or generating an image through $\mathcal{A}_{\mathrm{draw}}$:
\begin{equation*}
(r_t,a_t)\sim
\pi_\theta(\cdot\mid\mathcal{H}_{t-1})
\qquad
a_t\in
\mathcal{A}_{\mathrm{tools}}
\cup
\mathcal{A}_{\mathrm{draw}}.
\end{equation*}
The reasoning span identifies missing knowledge, assesses the currently available evidence, and determines whether the next step should acquire external information or generate an image.

\textbf{Tool calling} When $a_t\in\mathcal{A}_{\mathrm{tools}}$, the policy action is a command $a_t=(n_t,u_t)$, where $n_t\in \mathcal{T}$ specifies the tool and $u_t$ is the query. \texttt{TextSearch} retrieves factual knowledge, whereas \texttt{ImageSearch} retrieves visual references. The environment executes the command and returns a multimodal observation
\begin{equation*}
o_t
\sim
\mathcal{W}(\cdot\mid a_t).
\end{equation*}
The observation, including the returned text, images, and source-aware summaries, is appended to the history. Conditioned on this new evidence, the policy may continue searching, reformulate the query, switch tools, or proceed to drawing.

\textbf{Native Image Generation} When $a_t\in\mathcal{A}_{\mathrm{draw}}$, the policy action consists of a visual-caption span $g_t$ followed by a visual-token span $v_t$:
\begin{equation*}
a_t=(g_t,v_t).
\end{equation*}
The caption $g_t$ consolidates the user request and acquired evidence into an executable generation intent, while $v_t$ represents the generated image. Unlike a prompt-optimization agent, \ourmodel{} does not delegate this step to an external image generator: both $g_t$ and $v_t$ are produced by the UMM policy itself. A generated image remains in the multimodal history, so drawing does not necessarily terminate the interaction. The policy may inspect its current result, identify missing or incorrect content, invoke additional search tools, and generate a revised image.

\textbf{Agentic Trajectory} The outcome of round $t$ differs according to the selected action:
\begin{equation*}
h_t=
\begin{cases}
[r_t,a_t,o_t]
& a_t\in\mathcal{A}_{\mathrm{tools}},\\
[r_t,a_t]
& a_t\in\mathcal{A}_{\mathrm{draw}}.
\end{cases}
\end{equation*}
Starting from $\mathcal{H}_0=q$, the multimodal history is updated by $\mathcal{H}_t=[\mathcal{H}_{t-1},h_t]$. A complete trajectory with $T$ rounds is therefore
\begin{equation}
\mathcal{H}_T=[q,h_1,h_2,\ldots,h_T].
\label{eq:agentic_trajectory}
\end{equation}
The occurrence, ordering, and number of search and drawing actions are not specified by a fixed workflow. The policy determines the interaction path from the evolving context and terminates only when it decides that the latest drawing action constitutes the final answer. 

\textbf{Agentic Image Generation Masking} The serialized trajectory contains tokens generated autoregressively by the UMM together with the user input and environment observations inserted as context. Let $\mathbf{T}$ denote the collection of token instances belonging to the reasoning spans $r_t$ and action spans $a_t$. Let $\mathbf{O}$ contain the conditioning instruction $q$ and all environment-returned observations $o_t$:
\begin{equation*}
\mathbf{T}
=
\bigcup_t(r_t\cup a_t),
\qquad
\mathbf{O}
=
q\cup\bigcup_t o_t.
\end{equation*}
Here, the collections refer to token instances at their serialized positions rather than vocabulary identities. For the $j$-th serialized token $y_j$, the policy-support mask is
\begin{equation*}
M_j
=
\mathbb{I}[y_j\in\mathbf{T}].
\end{equation*}
Tokens in $\mathbf{O}$ remain available for conditional modeling but are excluded from the policy objective. Consequently, training directly optimizes the complete set of reasoning, searching, and drawing behaviors produced by the UMM policy.

\section{Training}

We post-train the UMM in two stages. The first stage is SFT, which itself consists of two data-processing steps: a teacher agent performs multi-turn rollouts with text search, image search, and an image-generation tool to produce raw interaction trajectories; a converter then removes the dependence on the image-generation tool and turns these trajectories into training samples for the UMM. In the second stage, RL, we use \textbf{Reason-Act-Draw GRPO (RAD-GRPO)} to directly optimize the complete policy through interaction with the open-world environment, using both intent-level and quality-level rewards.

\begin{figure}[t]
    \centering
    \includegraphics[width=\linewidth]{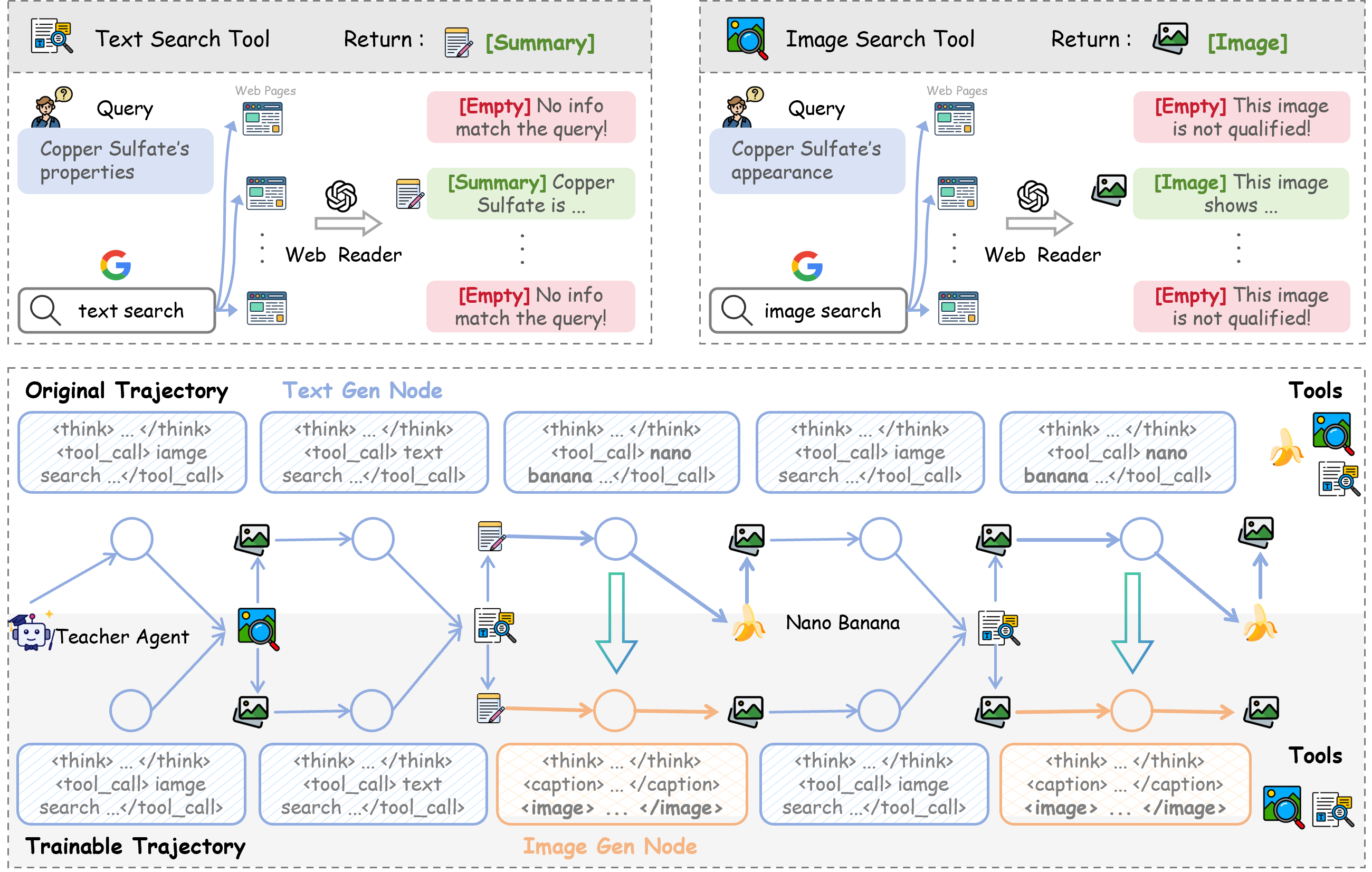}
    \caption{Post-training data and policy trajectory. The top part shows how text search and image search return evidence. The bottom part shows how raw trajectories are rewritten in the convert stage into native multimodal generation trajectories for SFT.}
    \label{fig:posttrain_pipeline}
\end{figure}

\subsection{Data Synthesis and Conversion}

As shown in Figure~\ref{fig:posttrain_pipeline}, when constructing the SFT data, the teacher agent follows user instructions through multi-turn interaction and calls \texttt{TextSearch}, \texttt{ImageSearch}, and an external image generator (we use \texttt{gemini-3-pro-image-preview}). The first two tools gather factual evidence and reference images, while the external image generator actually produces images during synthesis.

\paragraph{\textbf{Stage I: rollout.}}
The raw sample produced by the teacher agent is a complete multi-turn trajectory, including reasoning, tool calls, tool responses, and the final externally generated image. Unlike a static ``search first, draw later'' pipeline, the rollout here is alternating: the model can reason first, then call a tool, then keep reasoning or search again until it reaches the final generation intent.\\
The search tools are also enhanced for open-world image generation: \texttt{TextSearch} uses Google Search together with an LLM Reader to retrieve, summarize, and fall back across candidate pages; \texttt{ImageSearch} first filters out pages that cannot be downloaded, uses the LLM Reader to judge whether the content is relevant, and finally returns directly usable reference images together with structured summaries. This makes the evidence more stable and easier to use in later conversion and training.

\paragraph{\textbf{Stage II: convert.}}
In the convert stage, the raw trajectories are organized into unified token-level supervision samples. For the search part, the text and images returned by the tools are kept in the context. For the external image-generation step, the converter no longer treats it as a separate tool output; instead, it rewrites it into a native multimodal generation format: the prompt used to call the image-generation tool is written as a visual-caption span, followed by the corresponding image tokens. In other words, every external image generation in the teacher trajectory is converted into the form in which the model itself completes image generation in the same autoregressive trajectory. If a trajectory fails before the final image, or if image loading, tokenization, or context-length checks fail, the whole sample is filtered out.

\subsection{Supervised Fine-Tuning}

Let
\(
\mathcal{D}_{\mathrm{SFT}}
=
\{x_i,y_i\}_{i=1}^{N}
\)
denote the dataset obtained after rollout and convert, where
$x_i=q_i$ is the conditioning input and
\(y_i\) is the corresponding multimodal policy trajectory.
Each example is serialized as $\mathcal{H}_i=[x_i;y_i]=(u_{i,1},\ldots,u_{i,L_i})$, where \(L_i\) denotes the valid token-sequence length. Let $\mathcal T_i$ denote the set of policy-generated tokens in example $i$, as defined by the agentic image generation masking formulation in the preliminary section. Define the indicator
\begin{equation*}
M_{i,j}\triangleq\mathbb{I}\!\left[u_{i,j}\in\mathcal T_i\right].
\end{equation*}
The SFT objective is:
\begin{equation*}
\mathcal{L}_{\mathrm{SFT}}(\theta)
= -\mathbb{E}_{(x_i,y_i)\sim\mathcal{D}_{\mathrm{SFT}}}
\left[
\frac{
\sum_{j=1}^{L_i} M_{i,j}
\log P_\theta\left(u_{i,j}\mid u_{i,<j}\right)
}{
\sum_{j=1}^{L_i} M_{i,j}
}
\right].
\end{equation*}
This objective supervises only the textual and visual spans generated by the policy, while excluding the conditioning prefix and environment-returned observations. In other words, SFT does not learn a standalone search module or image generator; it learns a complete multimodal policy trajectory that first reasons, then calls tools, converts the acquired evidence into a final generation caption, and finally produces image tokens natively.

\subsection{Reason–Act–Draw GRPO}

SFT provides a stable initialization, but it remains limited by the coverage of the synthetic data and therefore struggles to discover better agentic image generation strategies. We further apply RAD-GRPO to directly optimize the complete policy on real inference trajectories.

For each input $q$, the rollout policy interacts with the open-world environment and samples a group of \(B\) trajectories:
\begin{equation*}
\mathcal{H}_i
\sim
\pi_{\theta_{\mathrm{old}}}(\cdot\mid q)
\otimes
\mathcal W
\qquad
i=1,\ldots,B.
\end{equation*}

Each trajectory may freely interleave reasoning, search, and visual generation. Let \(T_i\) denote the termination step of trajectory \(\mathcal{H}_i\). Its final visual-generation action is represented as $\left(g_{T_i},v_{T_i}\right)$, where \(g_{T_i}\) and \(v_{T_i}\) denote the final visual-caption span and the final visual-token span, respectively.

\paragraph{Dual Reward.}

RAD-GRPO does not only look at the final image. It looks at both the generation intent and the final image quality.

\begin{itemize}
    \item \textbf{Intent reward $R_i^{\mathrm{I}}$} This reward checks whether the final visual-caption span $g_{T_i}$ turns the user request and the acquired evidence into a sufficient, accurate, and executable generation description. In other words, it measures whether this prompt would be enough for an ideal generator to produce the right image. If the trajectory has no valid final caption, we set this term to 0.

    \item \textbf{Quality reward $R_i^{\mathrm{Q}}$} This reward checks whether the image decoded from $v_{T_i}$ really satisfies the user request and matches the final caption. In code, it is judged by a world-knowledge reward model with four dimensions: faithfulness, visual correctness, text accuracy, and aesthetics, combined with weights 0.1, 0.4, 0.4, and 0.1. If the task does not require readable text, text accuracy falls back to 0.5.
\end{itemize}

The two terms are combined into a main reward:
\begin{equation*}
R_i
=
\alpha R_i^{\mathrm{I}}
+
(1-\alpha)R_i^{\mathrm{Q}},
\qquad
\alpha\in[0,1].
\end{equation*}
The default is \(\alpha=0.5\). This means the caption term constrains whether the generation intent is good, while the image term constrains whether the final image is actually correct.

\paragraph{Auxiliary rewards.}

Besides the dual reward itself, RL also adds four auxiliary signals:
\begin{itemize}
    \item \textbf{Format reward} checks whether the output format is complete, whether the tool call is valid, and whether the think / caption / image structure matches the training format;
    \item \textbf{Draw signal} encourages the trajectory to actually reach native image generation;
    \item \textbf{Length penalty} discourages overly long trajectories;
    \item \textbf{No-draw penalty} gives an extra penalty if the trajectory never generates an image, so the policy does not collapse into a search-only mode.
\end{itemize}

Let $\widetilde R_i$ denote the final reward of trajectory $\mathcal H_i$, obtained by combining the dual reward with the format reward, draw signal, length penalty, and no-draw penalty. For each instruction $q$, we sample a group of $B$ trajectories from the old policy and optimize the following token-normalized GRPO objective:
\begin{equation}
\begin{aligned}
\mathcal{J}(\theta)
&=
\mathbb{E}_{\substack{
q\sim\mathcal D,
\{\mathcal H_i\}_{i=1}^{B}
\sim\pi_{\theta_{\mathrm{old}}}(\cdot\mid q)
}}
\\[-1mm]
&\quad
\Bigg[
\frac{1}{\sum_{i=1}^{B}\sum_{j=1}^{L_i}M_{i,j}}
\sum_{i=1}^{B}\sum_{j=1}^{L_i}M_{i,j}
\Bigg(
\min\bigg(
\rho_{i,j}(\theta)\widehat A_i,
\operatorname{clip}(\rho_{i,j}(\theta),
1-\epsilon,
1+\epsilon)\widehat A_i
\bigg)
-\beta\widehat D^{\mathrm{KL}}_{i,j}
\Bigg)
\Bigg].
\end{aligned}
\label{eq:rad_grpo}
\end{equation}
Here, $M_{i,j}$ is the policy-support mask defined in the preliminary section: it includes tokens generated in reasoning and action spans, while excluding the user instruction and environment-returned observations. The importance ratio and group-relative advantage are
\begin{equation}
\rho_{i,j}(\theta)
=
\frac{
\pi_\theta(u_{i,j}\mid u_{i,<j})
}{
\pi_{\theta_{\mathrm{old}}}(u_{i,j}\mid u_{i,<j})
},
\qquad
\widehat A_i
=
\frac{
\widetilde R_i-\operatorname{mean}(\{\widetilde R_b\}_{b=1}^{B})
}{
\operatorname{std}(\{\widetilde R_b\}_{b=1}^{B})+\varepsilon
}.
\label{eq:rad_grpo_ratio_advantage}
\end{equation}
The same trajectory-level advantage $\widehat A_i$ is applied to all policy-generated tokens in $\mathcal H_i$. Therefore, the final intent and image-quality feedback jointly optimize the complete sequence of reasoning, tool use, and native image generation. The KL term $\widehat D^{\mathrm{KL}}_{i,j}$ regularizes the updated policy toward the frozen SFT reference policy, and $\beta$ controls its strength.


\section{Experiments}

\subsection{Evaluation Setting}

\paragraph{\textbf{WISE.}}
WISE evaluates whether text-to-image models can integrate world knowledge rather than only perform shallow word-pixel alignment. It contains 1,000 prompts across 25 subdomains, grouped into cultural common sense, spatio-temporal reasoning, and natural science. In our table, we report its six category scores: Cultural, Time, Space, Biology, Physics, and Chemistry, together with Overall. The official WiScore uses an LLM-as-judge~\cite{li2024llmsasjudge} protocol over three criteria: image-text consistency, realism, and aesthetic quality, with weights 0.4, 0.3, and 0.3 respectively.

\paragraph{\textbf{WorldGenBench Humanities.}}
WorldGenBench targets reasoning-driven world-knowledge image generation. We use its Humanities split, which covers 244 countries and regions with 732 prompts, organized by continent. The benchmark constructs prompt-specific knowledge checklists and scores generated images by the Knowledge Checklist Score (KCS): each image is judged against the expected semantic score points in the checklist, such as culturally appropriate clothing, local architecture, region-specific tools, landmarks, livestock, vegetation, coastline, or readable labels. The final score is the normalized checklist satisfaction score. We report the continent-level scores AF, AN, AS, EU, NA, OC, and SA, as well as their average.

\begin{table*}[!t]
\centering
\caption{\textbf{Main results on WISE and WorldGenBench-Humanities.} Models are grouped into frontier proprietary models, general image-generation models, unified multimodal models, and agentic image-generation models. Dashes indicate unavailable entries in the collected source table. Lightly emphasized and \second{underlined} scores denote the best and second-best non-proprietary results within each metric column; ties share the same marker.}
\label{tab:main_results}
\setlength{\tabcolsep}{2.1pt}
\fontsize{8.2pt}{10.2pt}\selectfont
\renewcommand{\arraystretch}{1.16}
\begin{tabular}{>{\raggedright\arraybackslash}p{3.55cm} ccccccc cccccccc}
\toprule
\multirow{2}{*}{\textbf{Method}} &
\multicolumn{7}{c}{\textbf{WISE}} &
\multicolumn{8}{c}{\textbf{WorldGenBench-Humanities}} \\
\cmidrule(lr){2-8}\cmidrule(lr){9-16}
& Cul. & Time & Space & Bio. & Phys. & Chem & Avg. &
AF & AN & AS & EU & NA & OC & SA & Avg. \\
\midrule
\rowcolor[HTML]{f0f0f0}
\multicolumn{16}{c}{\textbf{\textit{Frontier Proprietary Models}}} \\
Nano Banana-Pro & 0.89 & 0.80 & 0.89 & 0.88 & 0.86 & 0.85 & 0.87 & 27.75 & 23.80 & 28.82 & 29.77 & 28.27 & 30.65 & 25.83 & 28.62 \\
Nano Banana & 0.89 & 0.87 & 0.95 & 0.89 & 0.89 & 0.79 & 0.89 & 29.91 & 17.69 & 31.10 & 28.93 & 29.41 & 31.37 & 26.53 & 29.67 \\
\midrule
\rowcolor[HTML]{f0f0f0}
\multicolumn{16}{c}{\textbf{\textit{General Image-Generation Models}}} \\
FLUX.1-schnell & 0.39 & 0.44 & 0.50 & 0.31 & 0.44 & 0.26 & 0.40 & 11.31 & 8.61 & 13.52 & 11.79 & 10.84 & 12.38 & 12.96 & 12.00 \\
SD-XL-base-0.9 & 0.43 & 0.48 & 0.47 & 0.44 & 0.45 & 0.27 & 0.43 & 10.89 & 9.09 & 11.47 & 10.14 & 9.94 & 9.54 & 10.74 & 10.55 \\
SD-3.5-medium & 0.43 & 0.50 & 0.52 & 0.41 & 0.53 & 0.33 & 0.45 & 12.08 & 11.94 & 12.44 & 11.40 & 10.15 & 13.33 & 12.69 & 11.85 \\
SD-3.5-large & 0.44 & 0.50 & 0.58 & 0.44 & 0.52 & 0.31 & 0.46 & 11.82 & 13.24 & 13.43 & 12.72 & 11.46 & 11.91 & 15.57 & 12.57 \\
PixArt-Alpha & 0.45 & 0.50 & 0.48 & 0.49 & 0.56 & 0.34 & 0.47 & 10.12 & 7.27 & 12.58 & 11.24 & 9.83 & 8.66 & 9.46 & 10.65 \\
Playground-v2.5 & 0.49 & 0.58 & 0.55 & 0.43 & 0.48 & 0.33 & 0.49 & 12.03 & 10.10 & 13.27 & 11.91 & 10.35 & 9.68 & 13.16 & 11.83 \\
FLUX.1-dev & 0.48 & 0.58 & 0.62 & 0.42 & 0.51 & 0.35 & 0.50 & 8.43 & 11.30 & 10.15 & 10.59 & 8.23 & 8.43 & 9.63 & 9.36 \\
LongCat-Image & 0.66 & 0.61 & 0.72 & 0.66 & 0.72 & 0.49 & 0.65 & 15.63 & \second{16.30} & 15.52 & 16.59 & \second{21.52} & 12.56 & \second{21.67} & 16.80 \\
\midrule
\rowcolor[HTML]{f0f0f0}
\multicolumn{16}{c}{\textbf{\textit{Unified Multimodal Models}}} \\
Janus-Pro-1B & 0.20 & 0.28 & 0.45 & 0.24 & 0.32 & 0.16 & 0.26 & 3.34 & 5.93 & 4.02 & 2.97 & 2.07 & 3.60 & 5.33 & 3.41 \\
VILA-u-7B-256 & 0.26 & 0.33 & 0.37 & 0.35 & 0.39 & 0.23 & 0.31 & 6.19 & 3.74 & 6.73 & 5.42 & 5.23 & 3.67 & 4.58 & 5.62 \\
Janus-Pro-7B & 0.30 & 0.37 & 0.49 & 0.36 & 0.42 & 0.26 & 0.35 & 6.87 & 5.45 & 8.57 & 9.22 & 5.28 & 5.24 & 8.46 & 7.41 \\
Emu3 & 0.34 & 0.45 & 0.48 & 0.41 & 0.45 & 0.27 & 0.39 & 10.44 & 9.35 & 11.85 & 12.57 & 9.84 & 10.00 & 11.77 & 11.13 \\
Emu3.5 & 0.73 & 0.39 & 0.33 & 0.25 & 0.36 & 0.33 & 0.49 & 13.19 & 9.44 & 12.77 & 14.75 & 6.98 & 18.45 & 17.78 & 13.17 \\
Hunyuan-Image-3.0 & 0.58 & 0.57 & 0.70 & 0.56 & 0.63 & 0.31 & 0.57 & 11.58 & 10.93 & 12.57 & 14.70 & 12.71 & 9.44 & 17.64 & 12.76 \\
BLIP3o-8B & 0.49 & 0.51 & 0.63 & 0.54 & 0.63 & 0.37 & 0.52 & 13.58 & 14.81 & 14.71 & 12.70 & 11.53 & 11.77 & \second{21.67} & 13.63 \\
BAGEL & 0.44 & 0.55 & 0.68 & 0.44 & 0.60 & 0.39 & 0.52 & 10.00 & 10.93 & 10.59 & 10.34 & 10.39 & 10.95 & 11.53 & 10.47 \\
BAGEL+CoT & 0.76 & 0.69 & 0.75 & 0.65 & \second{0.75} & 0.58 & 0.70 & 8.53 & 13.61 & 14.62 & 12.97 & 12.94 & 12.14 & 10.14 & 12.09 \\
\midrule
\rowcolor[HTML]{f0f0f0}
\multicolumn{16}{c}{\textbf{\textit{Agentic Image Generation Models}}} \\
Unify-Agent & \second{0.82} & {\fontfamily{bytesansmedium}\selectfont 0.75} & 0.74 & 0.72 & 0.73 & 0.70 & \second{0.77} & 13.88 & 13.80 & 15.43 & 15.65 & 15.61 & 14.78 & {\fontfamily{bytesansmedium}\selectfont 24.31} & 15.58 \\
GenSearcher-Qwen-Image & 0.80 & \second{0.71} & {\fontfamily{bytesansmedium}\selectfont 0.82} & {\fontfamily{bytesansmedium}\selectfont 0.76} & 0.74 & \second{0.75} & \second{0.77} & 12.40 & 9.35 & 18.46 & 14.71 & 10.33 & 10.16 & 12.64 & 13.66 \\
\hdashline
\rowcolor[HTML]{e8f0fe}
ToolArtist & {\fontfamily{bytesansmedium}\selectfont 0.87} & 0.62 & 0.75 & \second{0.75} & {\fontfamily{bytesansmedium}\selectfont 0.81} & {\fontfamily{bytesansmedium}\selectfont 0.79} & {\fontfamily{bytesansmedium}\selectfont 0.79} & {\fontfamily{bytesansmedium}\selectfont 24.44} & {\fontfamily{bytesansmedium}\selectfont 17.13} & {\fontfamily{bytesansmedium}\selectfont 26.19} & \second{18.69} & \second{19.52} & \second{20.97} & 19.44 & {\fontfamily{bytesansmedium}\selectfont 22.10} \\
\bottomrule
\end{tabular}
\end{table*}
\FloatBarrier

\paragraph{\textbf{Baselines.}}
We group baselines into four families. \textbf{\ding{182} Frontier Proprietary Models} include closed commercial image generators such as the Nano Banana series. \textbf{\ding{183} General Image Generation Models} include image generators without an explicit agent loop. \textbf{\ding{184} Unified Multimodal Models (UMMs)} are a distinct family that keeps image understanding and image generation inside one multimodal formulation rather than splitting them into separate planner and generator systems. \textbf{\ding{185} Agentic Image Generation Models.} These methods add reasoning, tool use, or prompt-level decomposition before synthesis, but we do not frame them primarily around search.

\subsection{Main Results}

\paragraph{\textbf{Result discussion.}}

On WISE, our model reaches 0.79 overall, outperforming the prior agentic image-generation models in the collected table. The strongest proprietary image models remain ahead on several WISE categories, especially Time and Space, but our method is competitive on knowledge-heavy natural-science categories, achieving 0.81 on Physics and 0.79 on Chemistry. On WorldGenBench-Humanities, our model obtains the best non-proprietary average KCS score, 22.10, compared with 21.76 for the strongest newly evaluated open-source baseline (Qwen-Image), 15.58 for Unify-Agent, and 13.66 for GenSearcher-Qwen-Image. The highest non-proprietary continent-level scores are split across methods: our model leads Africa, Antarctica, and Asia, Qwen-Image leads Europe, North America, and Oceania, and Unify-Agent leads South America.

\section{Analysis}

\subsection{RL Training Dynamics}
\begin{figure}[htbp]
    \centering
    \includegraphics[width=\textwidth]{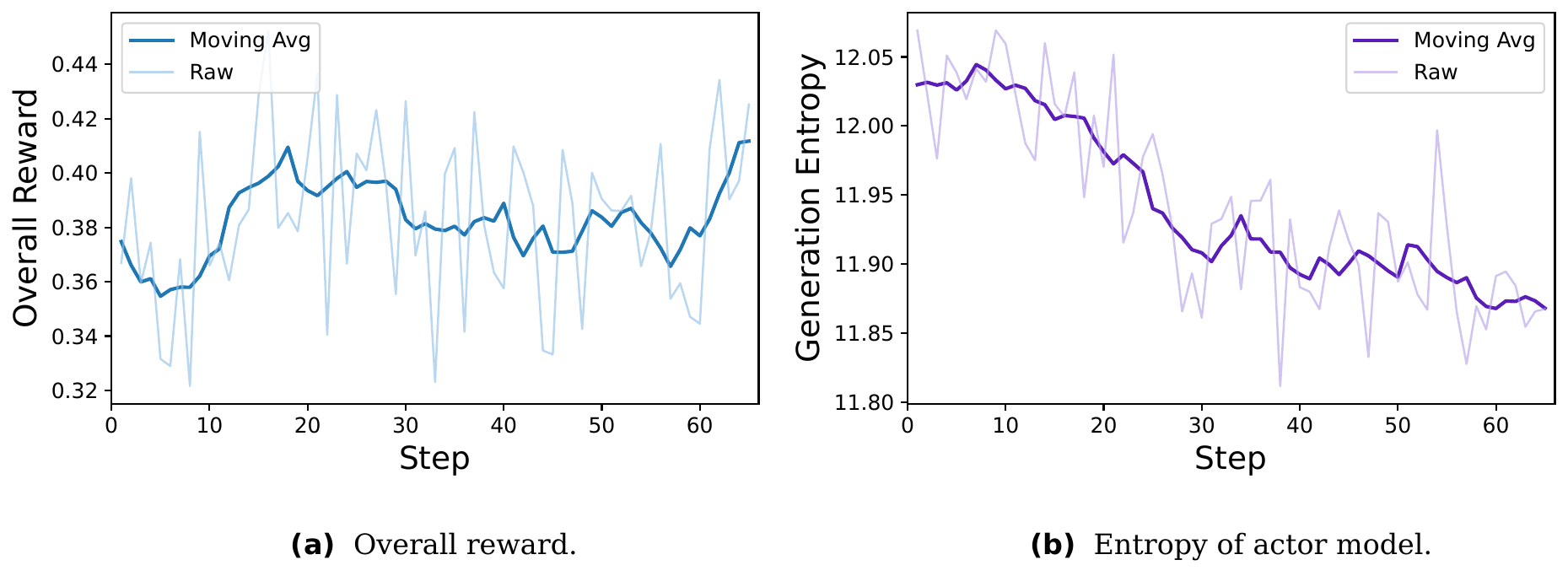}
    \caption{RAD-GRPO training dynamics.}
    \label{fig:rad_grpo_training_dynamics}
\end{figure}

\textbf{Reward Dynamics.} (a) shows the overall reward during RAD-GRPO training. Because the reward is jointly computed from online search trajectories, the Caption Reward, and the Image Reward, the raw reward exhibits noticeable fluctuations. Nevertheless, the moving-average curve shows an overall upward trend: after a brief decline at the beginning of training, the reward increases rapidly between approximately Steps 10 and 20; during the middle stage, it fluctuates between about 0.37 and 0.40; and toward the end of training, it rises further to approximately 0.41. This trend indicates that RAD-GRPO progressively improves the policy’s ability to generate high-quality search-to-image trajectories.

\textbf{Policy Entropy Dynamics.} (b) shows the generation entropy of the actor. At the beginning of training, the entropy remains at approximately 12.03; it then gradually decreases and stabilizes between about 11.87 and 11.91 during the latter half of training. The slight decrease in entropy accompanying the reward improvement suggests that the policy gradually concentrates probability mass on higher-reward reasoning, search, and generation behaviors. Notably, the entropy curve does not exhibit any sudden decline, and no evident entropy collapse is observed within the training process.

\subsection{Effect of Source-Aware Image Search Summaries}

The image search tool attaches an LLM-generated summary of the source webpage to each returned reference image. This design is intended to prevent the agent from selecting visually plausible but semantically wrong references. We ablate this component by removing image-source summaries while keeping the rest of the inference framework unchanged. As shown in \cref{tab:image_summary_ablation}, source-aware image summaries improve WISE Overall from 0.61 to 0.79. The largest drop appears in Biology, where the score decreases by 0.50, suggesting that image-source summaries are particularly important when the reference image must be tied to fine-grained factual identity rather than generic appearance.

\begin{table}[!htbp]
\centering
\caption{\textbf{Ablation of source-aware summaries in \texttt{image\_search}.} Red numbers denote drops from the full toolchain.}
\label{tab:image_summary_ablation}
\setlength{\tabcolsep}{0pt}
\normalsize
\renewcommand{\arraystretch}{1.12}
\newcommand{\abbase}[2]{#1}
\newcommand{\abdrop}[2]{#1\makebox[0pt][l]{\hspace{0.2em}{\scriptsize\bfseries\textcolor{red}{(#2)}}}}
\begin{tabular*}{\linewidth}{@{\extracolsep{\fill}}lccccccc@{\hspace{1.4em}}}
\toprule
\textbf{Setting} & \textbf{Cul.} & \textbf{Time} & \textbf{Space} & \textbf{Bio.} & \textbf{Phys.} & \textbf{Chem.} & \textbf{Avg.} \\
\midrule
\textbf{Full} &
\abbase{0.87}{-.09} &
\abbase{0.62}{-.08} &
\abbase{0.75}{-.21} &
\abbase{0.75}{-.50} &
\abbase{0.81}{-.17} &
\abbase{0.79}{-.29} &
\abbase{0.79}{-.18} \\
\textbf{w/o summary} &
\abdrop{0.78}{-.09} &
\abdrop{0.54}{-.08} &
\abdrop{0.54}{-.21} &
\abdrop{0.25}{-.50} &
\abdrop{0.64}{-.17} &
\abdrop{0.50}{-.29} &
\abdrop{0.61}{-.18} \\
\bottomrule
\end{tabular*}
\end{table}
\FloatBarrier

\subsection{Comparison}

\begin{figure}[!htbp]
\centering
\includegraphics[width=\linewidth]{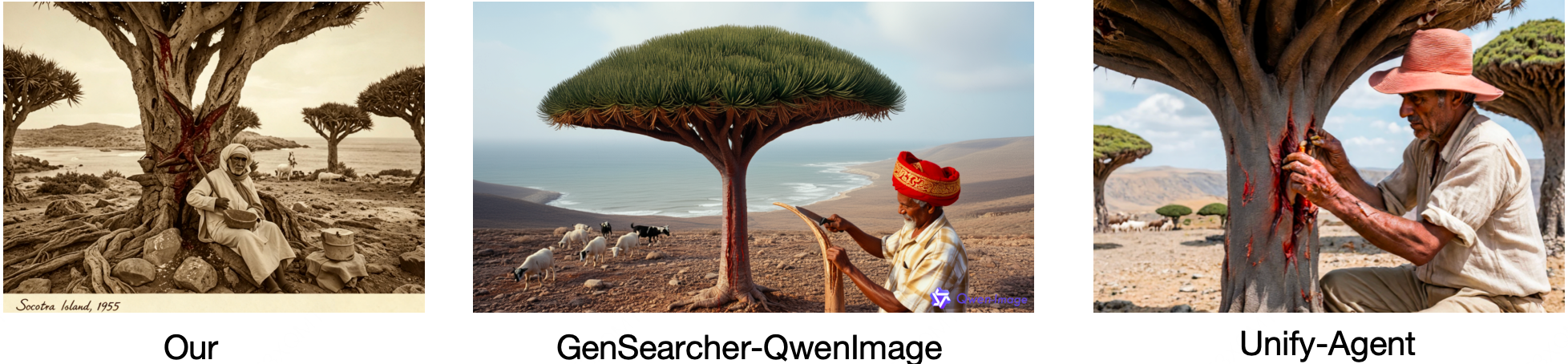}
\caption{A case on WorldGenBench-Humanities. Our model better grounds the Socotra resin-collection scene in traditional tools, clothing, storage containers, grazing goats, and coastline cues.}
\label{fig:worldgen_yemeni_case}
\end{figure}

\cref{fig:worldgen_yemeni_case} presents a WorldGenBench-Humanities case about dragon's-blood resin collection on Socotra Island in 1955. The full evaluation prompt is:

\begin{tcolorbox}[
colback=gray!3!white,
colframe=gray,
colbacktitle=gray,   
coltitle=white,               
fonttitle=\bfseries,          
title=WorldGenBench-Humanities Prompt,
left=1.5mm,
right=1.5mm,
top=1mm,
bottom=1mm]
\small\ttfamily
A dragon's blood tallow collector on Socotra Island is cleaning a split in the trunk in 1955. The sea breeze sends a salty scent, and in the distance the cries of sheep. These ancient trees have witnessed countless generations, and the collector's skills have been passed down from generation to generation.
\end{tcolorbox}

The prompt requires both cultural details and geographic grounding: a traditional resin collector, appropriate tools and clothing, dragon-blood trees, local livestock, storage containers, and the coastline of Socotra. Our RL checkpoint obtains KCS 0.625, while Unify-Agent obtains 0.000 and GenSearcher obtains 0.250. Our result satisfies five checklist points. First, the collector is shown using traditional resin-harvesting implements, including a wooden scraping tool and a clay-like container. Second, the figure wears loose traditional clothing and a head covering that fit the tropical island setting, rather than modern casual clothing. Third, grazing goats are visible in the scene, matching the local livestock requirement. Fourth, storage containers are placed near the collector, making the resin-collection activity concrete rather than only implied. Finally, the image includes a visible coastline and ocean, grounding the scene on Socotra Island. The remaining missed points are the limestone-mountain landscape, endemic island vegetation such as agave or desert thistle, and traditional stone houses.

\begin{table}[!htbp]
\centering
\caption{\textbf{Checklist evaluation for the Socotra case.} Each row is one score point from the benchmark checklist. Blue rows mark the points satisfied by our model. \ding{51} means satisfied and \ding{55} means not satisfied.}
\label{tab:worldgen_yemeni_checklist}
\scriptsize
\setlength{\tabcolsep}{5pt}
\renewcommand{\arraystretch}{1.12}
\resizebox{\linewidth}{!}{%
\begin{tabular}{lccc}
\toprule
Checklist item & Ours & Unify-Agent & GenSearcher \\
\midrule
\rowcolor[HTML]{e8f0fe}
Traditional resin-harvesting tools, such as wooden scrapers and clay pots & \ding{51} & \ding{55} & \ding{55} \\
\rowcolor[HTML]{e8f0fe}
Lightweight traditional clothing appropriate for the tropical climate & \ding{51} & \ding{55} & \ding{55} \\
Socotra landscape with limestone mountains & \ding{55} & \ding{55} & \ding{55} \\
\rowcolor[HTML]{e8f0fe}
Visible grazing goats, the main livestock in the area & \ding{51} & \ding{55} & \ding{51} \\
Typical island vegetation such as agave and desert thistle & \ding{55} & \ding{55} & \ding{55} \\
Traditional stone houses visible in the distance & \ding{55} & \ding{55} & \ding{55} \\
\rowcolor[HTML]{e8f0fe}
Storage containers for collected resin, possibly local-material products & \ding{51} & \ding{55} & \ding{55} \\
\rowcolor[HTML]{e8f0fe}
Visible coastline and Indian Ocean & \ding{51} & \ding{55} & \ding{51} \\
\midrule
KCS & 0.625 & 0.000 & 0.250 \\
\bottomrule
\end{tabular}
}
\end{table}
\FloatBarrier

Additional qualitative comparisons and their checklist-level evaluations are
provided in \cref{app:additional_comparisons}, covering a broader range of
historical, geographic, and cultural settings.

\subsection{SFT Data Distribution}

\begin{figure}[!htbp]
\centering
\includegraphics[width=\linewidth]{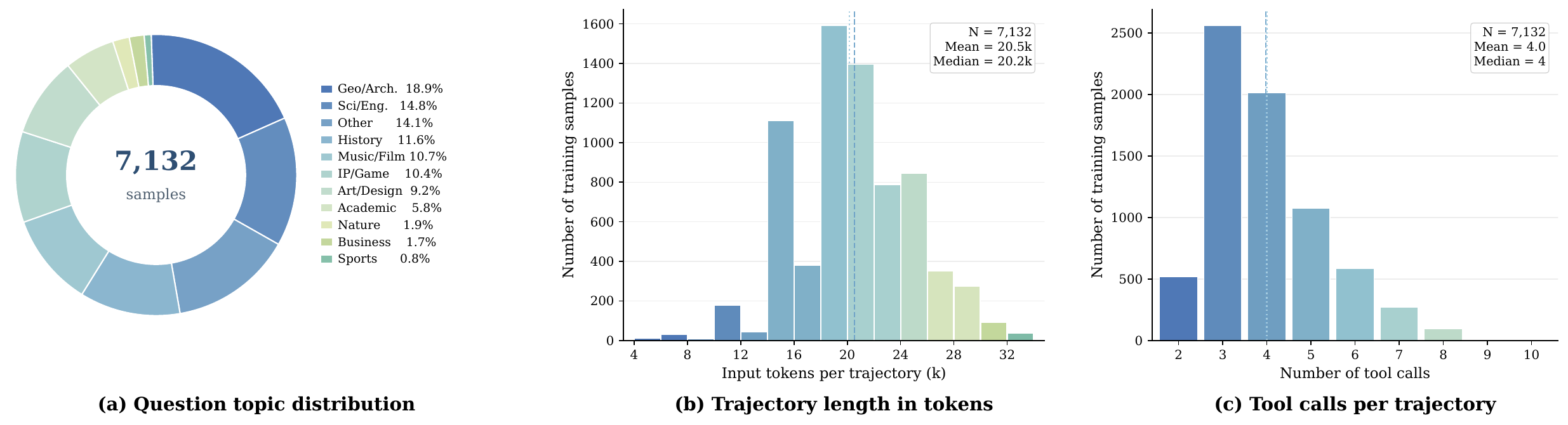}
\caption{Distribution of retained SFT trajectories. The figure reports topic categories, trajectory token length, and number of tool calls.}
\label{fig:sft_data_distribution}
\end{figure}

\cref{fig:sft_data_distribution} summarizes the retained SFT corpus. The final corpus contains 7,132 trajectories after conversion and filtering. The topic distribution is diverse: the largest category is Geo/Architecture (18.9\%), followed by Sci/Engineering (14.8\%), Other (14.1\%), History (11.6\%), Music/Film (10.7\%), IP/Game (10.4\%), Art/Design (9.2\%), Academic (5.8\%), Nature (1.9\%), Business (1.7\%), and Sports (0.8\%). Trajectory length is centered around long multimodal contexts, with a mean of 20.5k and a median of 20.2k input tokens. Tool use is also compact: the mean and median number of tool calls are both 4.0. This distribution suggests that the SFT corpus is not a collection of one-step image prompts; it mainly consists of multi-turn search-and-generation trajectories with substantial retrieved evidence and visual context.

\FloatBarrier

\subsection{Case Study}

\begin{tcolorbox}[
    promptbox,
    title={ReactGen Example: Thomas Edison},
    breakable,
    fontupper=\small\ttfamily,
    boxsep=2pt,
    top=3pt,
    bottom=3pt,
    left=5pt,
    right=5pt
]
\footnotesize
\setlength{\parindent}{0pt}
\setlength{\parskip}{1pt}

\textbf{Query:}

Thomas Edison's greatest invention.

\vspace{3pt}
\textbf{Step 1. Reasoning + Text Search:}

I first need to verify which invention is most widely regarded as
Thomas Edison's greatest contribution, ensuring factual accuracy for
subsequent image retrieval and generation.

\vspace{2pt}
\texttt{Thomas Edison most important invention}\\
\texttt{greatest invention of Thomas Edison}

\textbf{Summary:}

The practical incandescent light bulb is Edison's most iconic invention.
He improved earlier designs by developing a durable filament and the
electric power system required for practical use. The phonograph,
however, is often described as his first great invention and remained
his personal favorite.

\vspace{3pt}
\textbf{Step 2. Reasoning + Image Search:}

Because the phonograph is commonly described as Edison's first great
invention while the light bulb is his most famous contribution, the
final image should represent both interpretations.

\vspace{2pt}
\texttt{Thomas Edison official portrait}\\
\texttt{Edison phonograph original 1877 invention}\\
\texttt{Edison practical incandescent light bulb 1879}

\textbf{Summary:}

\vspace{2pt}
\noindent
\begin{minipage}[t]{0.30\linewidth}
    \centering
    \includegraphics[
        width=\linewidth,
        height=2.5cm,
        keepaspectratio
    ]{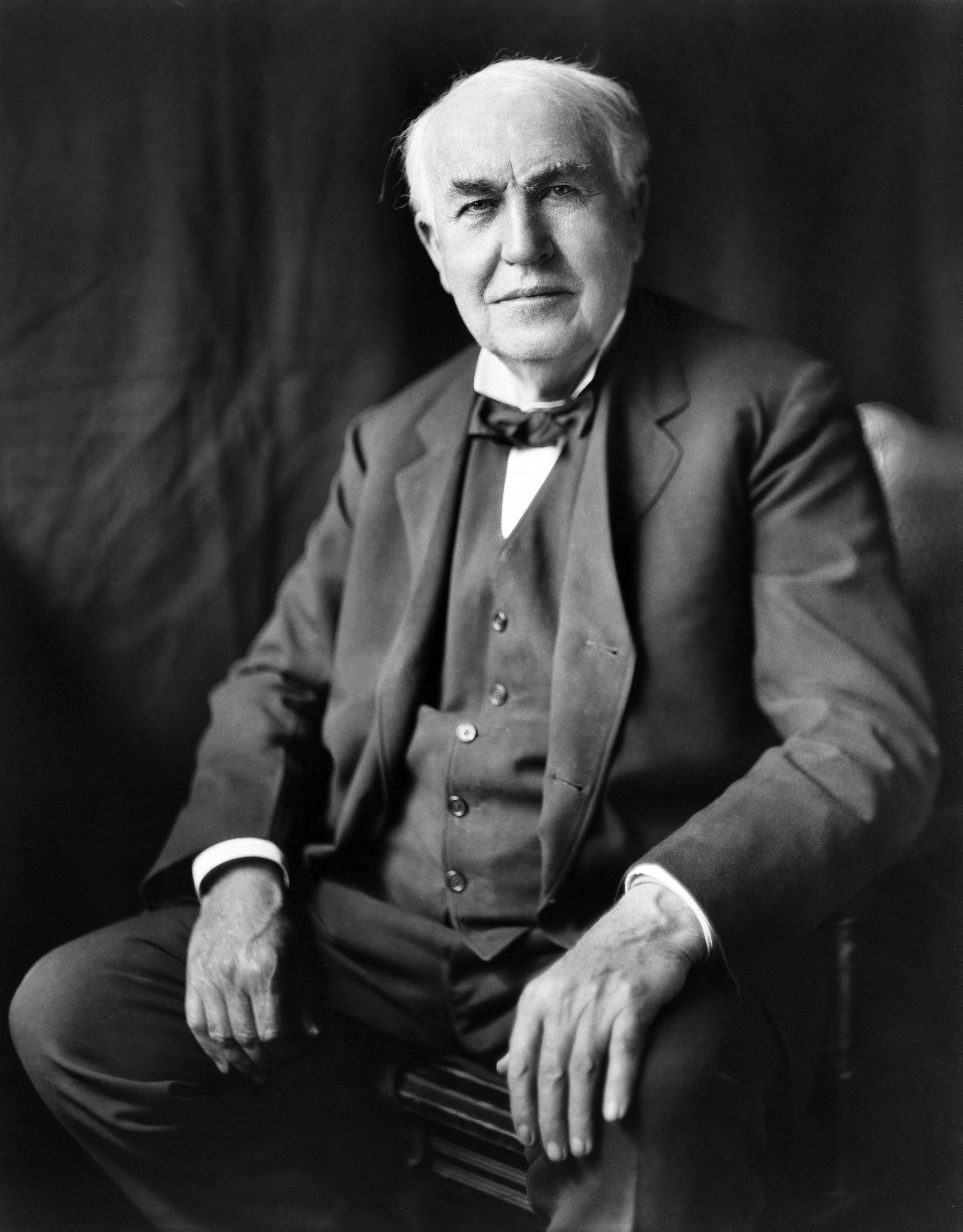}

    \vspace{1pt}
    \scriptsize\textbf{[IMAGE 1]}
\end{minipage}
\hfill
\begin{minipage}[t]{0.36\linewidth}
    \centering
    \includegraphics[
        width=\linewidth,
        height=2.5cm,
        keepaspectratio
    ]{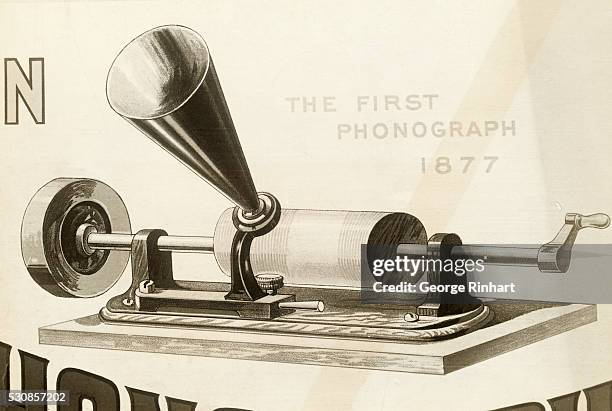}

    \vspace{1pt}
    \scriptsize\textbf{[IMAGE 2]}
\end{minipage}
\hfill
\begin{minipage}[t]{0.30\linewidth}
    \centering
    \includegraphics[
        width=\linewidth,
        height=2.5cm,
        keepaspectratio
    ]{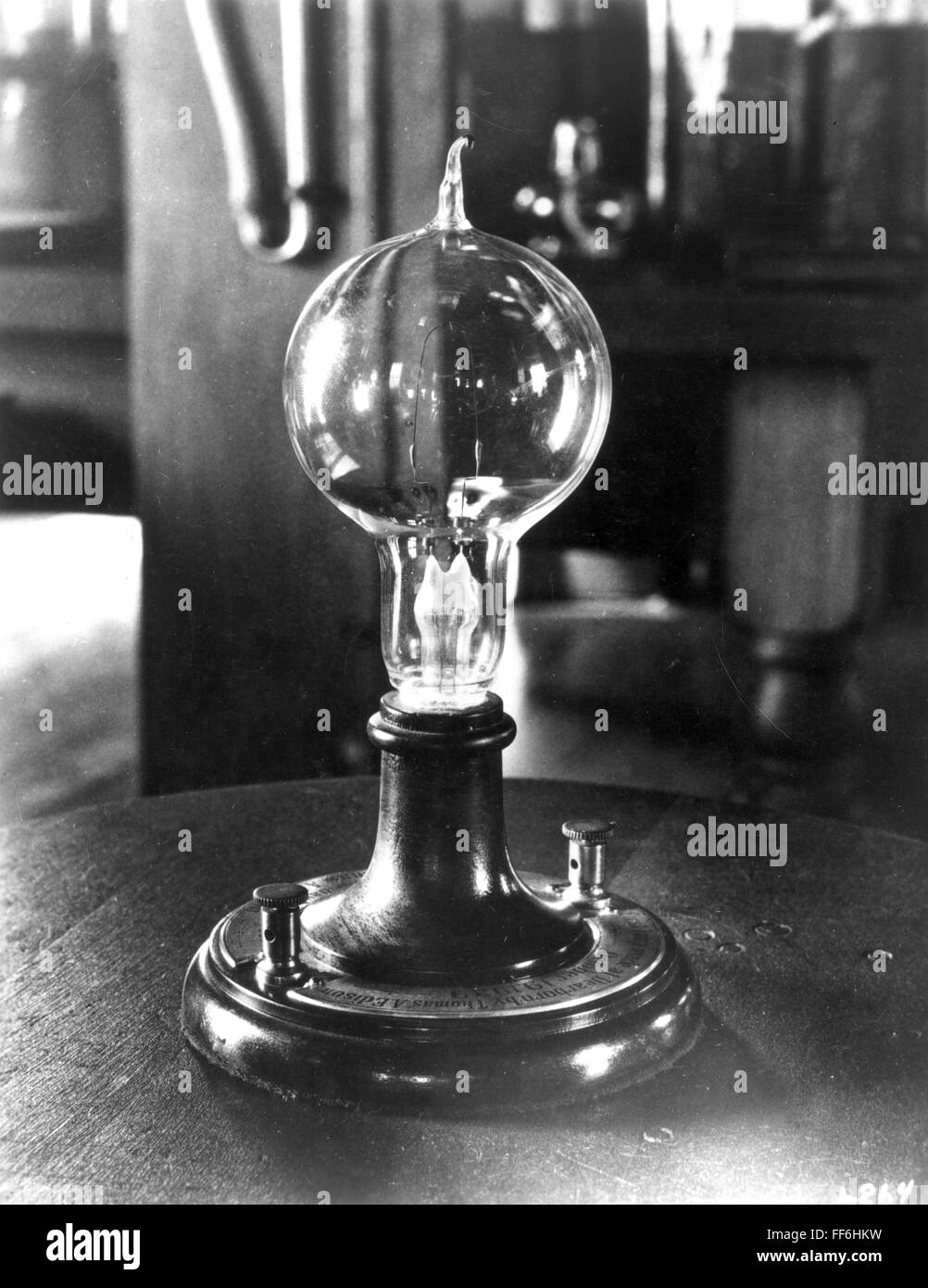}

    \vspace{1pt}
    \scriptsize\textbf{[IMAGE 3]}
\end{minipage}

\vspace{4pt}
\textbf{Step 3. Reasoning + Caption:}

To reflect both interpretations, the generated image should depict
Edison beside the phonograph and the practical incandescent light bulb
inside his laboratory.

\vspace{3pt}
\textbf{Step 4. Generating:}

\small
Vintage 1900s realistic historical photograph of Thomas Edison, use
\texttt{[IMAGE 1]} as reference for his exact facial features, clothing
and appearance, standing beside his two most iconic inventions: the
Edison phonograph, using \texttt{[IMAGE 2]} as reference for its
original design, on his left, and the practical incandescent light
bulb, using \texttt{[IMAGE 3]} as reference for its shape, on his
right, set in the interior of his West Orange laboratory, warm ambient
lighting, authentic 1900s photo aesthetic, high detail, 4K resolution.

\begin{center}
    \includegraphics[
        width=0.26\linewidth,
        keepaspectratio
    ]{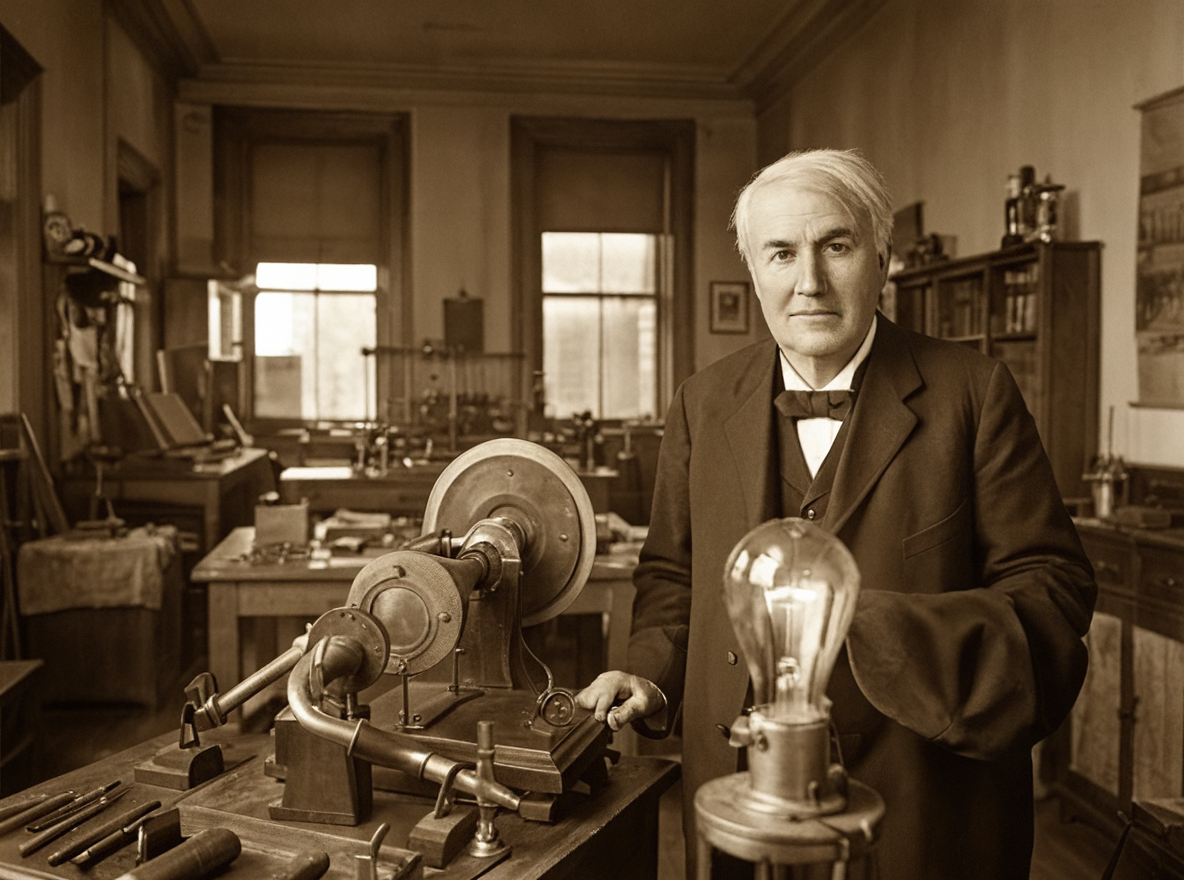}
\end{center}

\end{tcolorbox}

More complete agent trajectories, including intermediate reasoning, search
queries, retrieved multimodal evidence, the final generation caption, and the
resulting image, are presented in \cref{app:additional_trajectories}.

\section{Related Work}

\textbf{Unified Multimodal Model.}
Text-to-image models based on specialized diffusion or flow-based generators have achieved strong visual quality~\cite{podell2023sdxl,chen2023pixart,li2024playgroundv25,greenberg2025flux,meituan2025longcatimage}, but they do not natively express reasoning, multimodal observations, and image synthesis as actions of one policy. Unified multimodal models (UMMs) instead bring understanding and generation into a shared model. Janus-Pro separates the visual pathways for understanding and generation while retaining a shared autoregressive backbone~\cite{chen2025janus}; BLIP3-o generates semantic image features with a diffusion transformer~\cite{chen2025blip3o}; and BAGEL learns multimodal understanding, generation, and manipulation from large-scale interleaved pretraining~\cite{deng2025bagel}. Most relevant to our work is the Emu line. Emu3 tokenizes text, images, and video into discrete sequences and models all modalities solely by next-token prediction~\cite{wang2024emu3}. Emu3.5 scales this native autoregressive formulation to interleaved vision-language inputs and outputs and further strengthens reasoning and generation through post-training~\cite{cui2025emu35}. ToolArtist is built by post-training Emu3.5: we turn its native text-and-image token policy into an open-world agent that can decide when to reason, search, and draw.

\textbf{Tool-Using Agents and Agentic Image Generation.}
ReAct establishes the general pattern of interleaving reasoning with environment actions~\cite{yao2023react}. In image generation, retrieval-augmented methods first condition synthesis on external evidence~\cite{chen2022reimagen}, while later systems add planning, iterative refinement, or tool orchestration~\cite{wu2023visualchatgpt,yang2024idea2img,shalevarkushin2026imagerag,ye2025genpilot,jiang2026genagent,chen2026genevolve,ye2026genclaw,zhang2026qwenimageagent}. Unify-Agent uses a UMM in a structured pipeline of prompt understanding, multimodal evidence search, grounded recaptioning, and final synthesis~\cite{chen2026unifyagent}. GenSearcher trains a search agent with SFT and RL to perform multi-hop textual and visual retrieval, then supplies the grounded result to a separate image generator~\cite{feng2026gensearcher}. SearchGen further shows that search should be invoked selectively according to the generator's evolving knowledge boundary, since indiscriminate retrieval can introduce noise~\cite{wang2026searchgen}. These works motivate external knowledge acquisition, but differ in where agency resides: generation is either a prescribed stage or is delegated downstream after the learned search policy finishes.

\textbf{Reinforcement Learning for Visual Generation.}
GRPO removes the learned critic and estimates relative advantages from groups of sampled outputs, providing an efficient basis for outcome-driven reasoning optimization~\cite{shao2024deepseekmath}. Recent work extends group-relative optimization to unified visual generation. UniGRPO jointly optimizes reasoning and image synthesis for a single reasoning-then-generation round~\cite{liu2026unigrpo}, while interleaved GRPO studies unified optimization over multimodal text--image outputs~\cite{nie2026interleavedgrpo}.

\newpage

\bibliography{ref}
\bibliographystyle{plainnat}


\newpage
\appendix
\onecolumn

\section{Appendix}
\label{app:supplementary}

This appendix provides additional implementation details and qualitative
evidence for the post-training procedure and analyses presented in the main
paper. \cref{app:prompt_displays} reports the complete prompts used by the
teacher agent during data synthesis and by ToolArtist at inference time,
including the tool definitions, interaction protocol, and native image-
generation format. \cref{app:additional_comparisons} extends the main-text
comparison with four additional WorldGenBench-Humanities cases spanning
different historical, geographic, and cultural settings. For each case, we
show outputs from ToolArtist, Unify-Agent, and GenSearcher together with the
prompt-specific knowledge checklist, making both the improvements and the
remaining failure modes explicit. \cref{app:additional_trajectories} provides
a complete ToolArtist trajectory, exposing the intermediate reasoning, search
queries, retrieved textual and visual evidence, final visual caption, and
natively generated image. Finally, \cref{app:sft_examples} presents raw SFT
rollouts with their reference images and generated outputs, illustrating how
the synthesized teacher trajectories are represented as multimodal supervision
for post-training.

\subsection{Prompt Displays}
\label{app:prompt_displays}

\lstdefinestyle{fullpromptlisting}{
    basicstyle=\small\ttfamily,
    columns=fullflexible,
    breaklines=true,
    breakatwhitespace=false,
    keepspaces=true,
    showstringspaces=false,
    tabsize=2,
    literate={→}{{$\to$}}1 {—}{{--}}1
             {text_search}{\textbf{\textcolor{red}{text\_search}}}{10}
             {image_search}{\textbf{\textcolor{red}{image\_search}}}{11}
             {draw}{\textbf{\textcolor{red}{draw}}}{4}
}

\begin{tcblisting}{
    promptbox,
    title={Teacher Agent Full Prompt},
    breakable,
    listing only,
    listing options={style=fullpromptlisting},
    boxsep=2pt,
    top=3pt,
    bottom=3pt,
    left=5pt,
    right=5pt
}
SYSTEM_PROMPT = """You are a "retrieval-augmented image generation Agent". Your task is to proactively search for web information and reference images based on the user's image generation request, compile a high-quality image generation prompt, and then call the image generation tool to produce the image.

Action principles:
- For uncertain details involving real people, places, logos, clothing, architecture, IP styles, etc., prioritize using text_search to obtain textual descriptions, then use image_search to fetch reference images.
- Images returned by image_search will be shown to you directly in a multimodal way. Judge quality and relevance based on the image content, and pick the most suitable ones as references.
- Before calling draw, distill the retrieved information into a clear visual description: subject, scene, style, composition, color, material, action, lighting. Do not dump raw search results into the prompt.
- Generated images may not be perfect. After generation, carefully review the returned image. If unsatisfied, adjust the prompt or swap/add/remove reference images, and call draw multiple times to iteratively refine until the result is satisfactory. Do not stop after a single generation.
- Finally, wrap the answer with <|box_start|> and <|box_end|>, clearly stating the saved path of the final generated image or the reason for failure.

Reference image citation format (strict):
- When referring to reference images in the prompt field of draw, you can **only** use fixed tokens: [IMAGE1], [IMAGE2], [IMAGE3], ... (1-indexed, corresponding to the order of the images array).
- It is strictly forbidden to use natural language expressions such as "Image 1", "the first image", "the 1st image", "the first reference image", etc.
- Example: 'Paste the character from [IMAGE1] onto the background of [IMAGE2]'."""


USER_PROMPT = """A conversation between User and Assistant. The user provides an image generation request. The assistant searches for information and reference images, then generates the image by calling tools.

<tools>
{
  "name": "text_search",
  "description": "Batch web search tool. Takes an array of queries; for each query, uses Google search to obtain candidate pages and uses Jina Reader to fetch the title and body Markdown of the rank-1 page (very long content is truncated and marked with a truncated flag).",
  "parameters": {
    "type": "object",
    "properties": {
      "query": {"type": "array", "items": {"type": "string"}, "description": "Array of search queries; you can pass multiple complementary queries at once"}
    },
    "required": ["query"]
  }
}
{
  "name": "image_search",
  "description": "Batch image search tool. Takes an array of queries and returns structured information such as image titles, image URLs, width and height. Images are also shown to you in a multimodal way. The returned results[i].url can be directly placed into draw's images array.",
  "parameters": {
    "type": "object",
    "properties": {
      "query": {"type": "array", "items": {"type": "string"}, "description": "Array of image search queries; you can pass multiple complementary queries at once"}
    },
    "required": ["query"]
  }
}
{
  "name": "draw",
  "description": "Generates an image from a prompt, optionally with several reference images. The images array corresponds to [IMAGE1], [IMAGE2], ... in order; each item can be a local file path or a remote URL, and the tool will distinguish between them internally. Returns the local path where the image was saved and shows the generated image back to you in a multimodal way.",
  "parameters": {
    "type": "object",
    "properties": {
      "prompt": {"type": "string", "description": "Image generation prompt. References to reference images must use fixed tokens like [IMAGE1], [IMAGE2], etc.; any natural-language reference is forbidden"},
      "images": {"type": "array", "items": {"type": "string"}, "description": "Reference images, corresponding to [IMAGEk] in order. A single image should also be written as an array. Each item is either a local file path or a remote URL"}
    },
    "required": ["prompt"]
  }
}
</tools>

Workflow: think → call tool → wait for response → continue thinking or give an answer. Multiple rounds of calls are allowed.
Important: After generating an image, carefully review the returned image. If unsatisfied, adjust the prompt or reference images and call draw again, iterating until satisfied.

Example flow structure (only shows the call sequence; specific content should be filled in based on actual needs):

Analyze the user's request and plan what information and images to search for
<tool_call>
{"name": "text_search", "arguments": {"query": ["keyword1", "keyword2"]}}
</tool_call>
<tool_response>Search results</tool_response>
Extract useful information and decide which reference images to search for
<tool_call>
{"name": "image_search", "arguments": {"query": ["image keyword1", "image keyword2"]}}
</tool_call>
<tool_response>Image search results (you can see the images directly)</tool_response>
Pick suitable reference images based on what you see, integrate the information and write the image generation prompt
<tool_call>
{"name": "draw", "arguments": {"prompt": "..., referencing the styling of [IMAGE1], placing it into the scene of [IMAGE2]", "images": ["https://example.com/ref1.jpg", "https://example.com/ref2.jpg"]}}
</tool_call>
<tool_response>Generation result (you can see the generated image directly)</tool_response>
Check the generation result. If unsatisfied, adjust the prompt and call draw again; if satisfied, output the final answer
<|box_start|>Final result description (including the saved path of the generated image)<|box_end|>

Notes:
- Each round outputs only one tool_call or the final box answer
- tool_call must be valid JSON, including name and arguments
- References to reference images in draw's prompt can only use fixed tokens like [IMAGE1], [IMAGE2], ...; natural-language references such as "the first image" or "Image 1" are forbidden

User: """
\end{tcblisting}
\vspace{1cm}
\begin{tcblisting}{
    promptbox,
    title={ToolArtist Full Prompt},
    breakable,
    listing only,
    listing options={style=fullpromptlisting},
    boxsep=2pt,
    top=3pt,
    bottom=3pt,
    left=5pt,
    right=5pt
}
SYSTEM_PROMPT = """You are a "retrieval-augmented image-generation agent". Given the user's image-generation request, you actively search the web for information and reference images, then generate a high-quality image yourself.

Operating principles:
- For uncertain details about real people, places, logos, clothing, architecture, IP styles, etc., use text_search first to gather textual facts, then image_search to pull reference images.
- Images returned by image_search are shown to you directly. Judge their quality and relevance from what you see, and pick the most suitable ones as references.
- Before generating, distill the retrieved information into a clear visual description: subject, scene, style, composition, color, material, action, lighting. Do NOT dump raw search results into the pre-generation caption.
- The generated image may not be perfect. After generation, inspect it carefully; if unsatisfied, adjust the caption or swap/add/remove references and iterate until the result is good.

Reference-image citation format (strict):
- In the pre-generation caption you may ONLY refer to images via the fixed tokens [IMAGE 1], [IMAGE 2], [IMAGE 3], ... (1-indexed, matching the global numbering you see in tool responses).
- Natural-language references such as "Image 1", "the first image", "the first reference image", etc. are forbidden.
- Example: 'place the character from [IMAGE 1] onto the background of [IMAGE 2]'.

The user provides an image-generation request. You will search for information and reference images by calling tools, and then generate the image yourself.

<tools>
{
  "name": "text_search",
  "description": "Batch web search. Takes an array of queries; for each query, runs a Google search to find candidate pages and uses Jina Reader to fetch the rank-1 page's title and Markdown body (long pages are truncated with a `truncated` marker).",
  "parameters": {
    "type": "object",
    "properties": {
      "query": {"type": "array", "items": {"type": "string"}, "description": "Array of search queries; you may pass multiple complementary queries at once."}
    },
    "required": ["query"]
  }
}
{
  "name": "image_search",
  "description": "Batch image search. Takes an array of queries; returns image titles, URLs, sizes, and short summaries. Each image is also rendered to you directly. Every returned image is assigned a global [IMAGE n] index that you must use when referring to it later in the pre-generation caption.",
  "parameters": {
    "type": "object",
    "properties": {
      "query": {"type": "array", "items": {"type": "string"}, "description": "Array of image-search queries; you may pass multiple complementary queries at once."}
    },
    "required": ["query"]
  }
}
</tools>

Workflow: think → call a tool → wait for response → keep thinking or proceed to image generation. Multiple rounds allowed.

How to generate an image:
- When you are ready to generate, write a single pre-generation caption inside <|extra_50|> ... <|extra_51|>. This caption is the full visual description of the image you are about to produce: subject, scene, style, composition, color, material, action, lighting.
- Inside the caption, reference selected reference images by their global [IMAGE n] tokens (the same n that appeared in the tool responses). Only include references you actually want the model to use; omit unused images entirely.
- Immediately after closing the caption with <|extra_51|>, emit the image tokens for the generated image.

Example flow (structure only; fill in real content per request):

<|extra_60|>Analyse the user's request and plan what information and reference images are needed.<|extra_61|>
<tool_call>
{"name": "text_search", "arguments": {"query": ["keyword 1", "image keyword 2"]}}
</tool_call>
<tool_response>search results</tool_response>
<|extra_60|>Extract useful facts and decide which reference images to search.<|extra_61|>
<tool_call>
{"name": "image_search", "arguments": {"query": ["image keyword 1", "image keyword 2"]}}
</tool_call>
<tool_response>image results, each tagged with a global [IMAGE n] index and rendered visually</tool_response>
<|extra_50|>Combine the retrieved information and chosen references into the final visual description, citing only the [IMAGE n] tokens you want to use.<|extra_51|>
(generated image tokens)

Notes:
- Each tool_call must be valid JSON containing `name` and `arguments`.
- In the <|extra_50|>...<|extra_51|> caption, the ONLY allowed way to refer to a reference image is the fixed token [IMAGE 1], [IMAGE 2], ... -- never "Image 1", "the first image", "the first reference image", etc.
- Use the same global [IMAGE n] numbering that appears in the tool responses; do not renumber."""

USER_PREFIX = "User: "
\end{tcblisting}
\vspace{1cm}
\begin{tcolorbox}[
    promptbox,
    title={LLM Reader Prompt},
    breakable,
    boxsep=2pt,
    top=3pt,
    bottom=3pt,
    left=5pt,
    right=5pt,
    fontupper=\footnotesize\fontfamily{cmtt}\selectfont
]
\setlength{\parindent}{0pt}
\setlength{\parskip}{2pt}

You are an information extraction assistant. Below is a query and a piece of
document content. Find the most relevant core information from the document
related to the query, condense and organize it, with the following requirements:

\begin{enumerate}[leftmargin=14pt,itemsep=1pt,topsep=1pt]
    \item Only keep facts, data, or opinions that are truly relevant to the
    query; remove unrelated content.
    \item Present the information in compact bullet points or short paragraphs.
    Do not use pleasantries, restate the query, or include filler phrases such
    as ``according to the document''.
    \item If the document contains no relevant information at all, output only
    \texttt{NO\_RELEVANT\_INFO}.
    \item Preserve specific information from the original text, such as key
    numbers, names, times, and sources. Avoid being overly vague.
    \item Answer in the same language as the query.
\end{enumerate}

\textbf{Template fields:} \texttt{[query] \{query\}} and
\texttt{[document content] \{doc\}}. The reader directly outputs the condensed
and organized information.
\end{tcolorbox}

\subsection{Comparisons}
\label{app:additional_comparisons}

The following four cases are selected from the top-ranked
WorldGenBench-Humanities examples where our RL checkpoint outperforms both
Unify-Agent and GenSearcher. Each figure compares Ours, Unify-Agent, and
GenSearcher in that order.

\begin{figure}[!htbp]
\centering
\includegraphics[width=\linewidth]{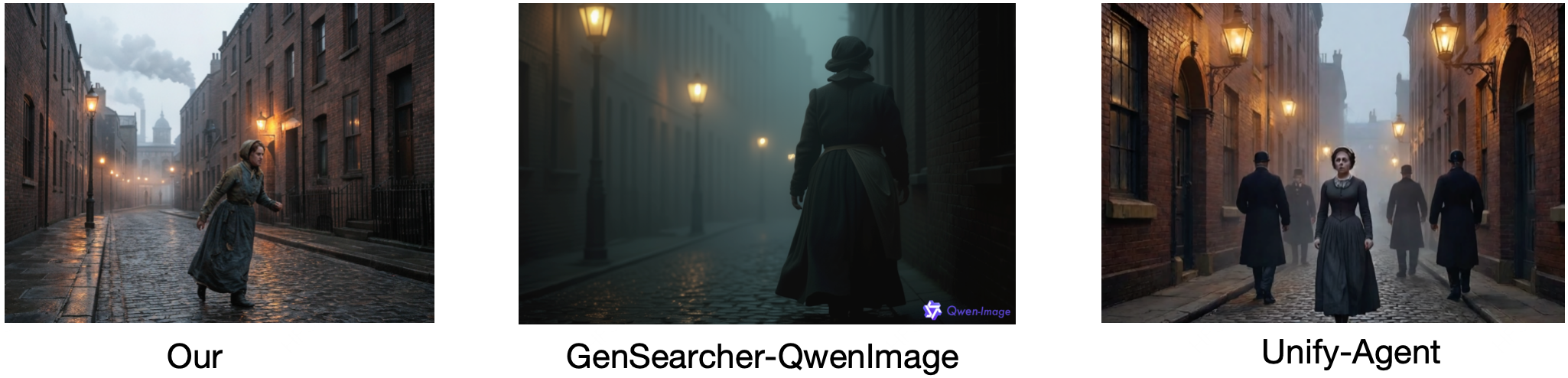}
\caption{Case study on WorldGenBench-Humanities. Our model better grounds the
Whitechapel scene in patched working-class dress, coal-smog atmosphere, and
soot-darkened Victorian facades while preserving the shared street, housing,
and gaslight cues.}
\label{fig:worldgen_whitechapel_case}
\end{figure}

\cref{fig:worldgen_whitechapel_case} presents a WorldGenBench-Humanities case
about a working woman in Whitechapel, East London, in 1889. The full evaluation
prompt is:

\begin{tcolorbox}[
colback=gray!3!white,
colframe=gray,
colbacktitle=gray,   
coltitle=white,               
fonttitle=\bfseries,          
title=WorldGenBench-Humanities Prompt,
left=1.5mm,
right=1.5mm,
top=1mm,
bottom=1mm]
\small\ttfamily
On an overcast November day in 1889, a working woman in the Whitechapel
district of East London was walking home. A thick fog shrouded the narrow
streets, and the gas streetlights glowed dimly. She picks up her pace as she
walks past a crowded room of low-cost apartments, and the smell of factory
smoke fills the air. It had been only a year since the Jack the Ripper case,
and the neighborhood was still shrouded in fear.
\end{tcolorbox}

The prompt requires both Victorian East End social cues and atmospheric
grounding: cobblestone streets, patched working-class dress, brick townhouse
apartments, cast-iron gas streetlights, coal smog, street occupations, police
patrols, soot-darkened facades, street sanitation details, and faded curtains.
Our RL checkpoint obtains KCS 0.600, while Unify-Agent obtains 0.300 and
GenSearcher obtains 0.300. Our result satisfies six checklist points. First,
the road is paved with round cobblestones. Second, the woman wears a patched
dark dress, capturing the working-class poverty cue. Third, the street is
lined with brick Victorian townhouse apartments. Fourth, the gas streetlights
use the expected cast-iron posts and glass shades. Fifth, tawny coal smoke is
visible in the air. Sixth, soot-darkened facades make the industrial pollution
concrete. The remaining missed points are newspaper boys, police patrols with
sirens and batons, horse manure and garbage, and faded curtains.

\begin{table}[!htbp]
\centering
\caption{\textbf{Checklist evaluation for the Whitechapel case.} Each row is
one score point from the benchmark checklist. Blue rows mark the points
satisfied by our model. \ding{51} means satisfied and \ding{55} means not
satisfied.}
\label{tab:worldgen_whitechapel_checklist}
\scriptsize
\setlength{\tabcolsep}{5pt}
\renewcommand{\arraystretch}{1.12}
\resizebox{\linewidth}{!}{%
\begin{tabular}{lccc}
\toprule
Checklist item & Ours & Unify-Agent & GenSearcher \\
\midrule
\rowcolor[HTML]{e8f0fe}
The streets are paved with round cobblestones & \ding{51} & \ding{51} & \ding{51} \\
\rowcolor[HTML]{e8f0fe}
Female workers wear patched and repaired dark-colored dresses & \ding{51} & \ding{55} & \ding{55} \\
\rowcolor[HTML]{e8f0fe}
The street is lined with brick Victorian townhouse apartments & \ding{51} & \ding{51} & \ding{51} \\
\rowcolor[HTML]{e8f0fe}
The gas streetlight is cast iron with a glass shade on top & \ding{51} & \ding{51} & \ding{51} \\
\rowcolor[HTML]{e8f0fe}
Tawny smoke floats in the air & \ding{51} & \ding{55} & \ding{55} \\
Newspaper boys selling newspapers on the street & \ding{55} & \ding{55} & \ding{55} \\
Visible police patrols equipped with sirens and batons & \ding{55} & \ding{55} & \ding{55} \\
\rowcolor[HTML]{e8f0fe}
Building facades blackened by soot & \ding{51} & \ding{55} & \ding{55} \\
Horse manure and garbage in the streets & \ding{55} & \ding{55} & \ding{55} \\
Faded curtains hang from the windows & \ding{55} & \ding{55} & \ding{55} \\
\midrule
KCS & 0.600 & 0.300 & 0.300 \\
\bottomrule
\end{tabular}
}
\end{table}
\FloatBarrier

\begin{figure}[!htbp]
\centering
\includegraphics[width=\linewidth]{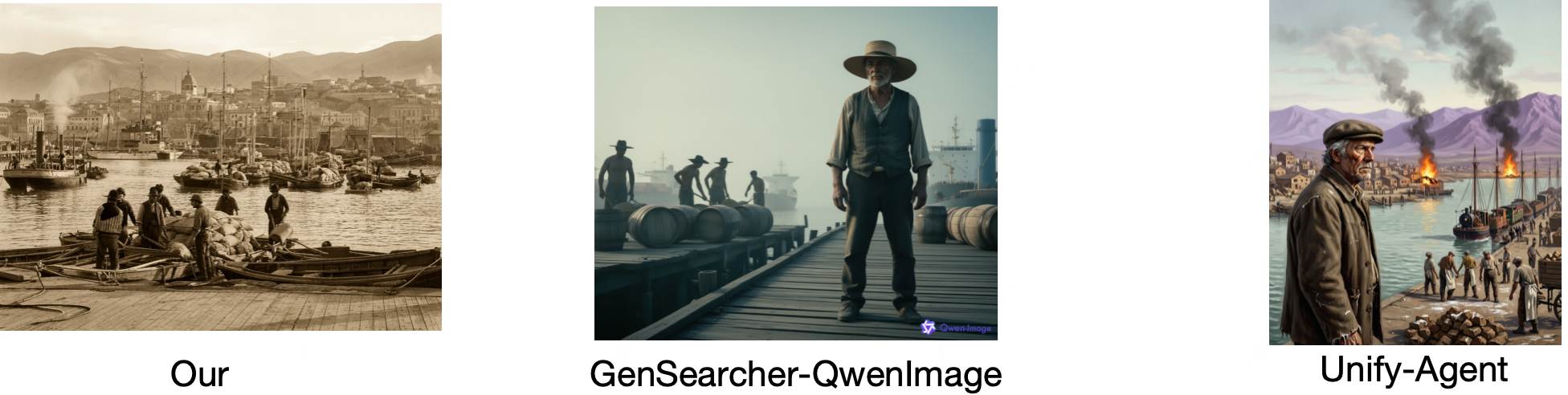}
\caption{Case study on WorldGenBench-Humanities. Our model better grounds the
Antofagasta port scene in period labor clothing, colonial masonry harbor
buildings, a distant steam freighter, desert surroundings, and morning fog.}
\label{fig:worldgen_antofagasta_case}
\end{figure}

\cref{fig:worldgen_antofagasta_case} presents a WorldGenBench-Humanities case
about a saltpeter miner at the port of Antofagasta during the Chilean Civil
War in February 1891. The full evaluation prompt is:

\begin{tcolorbox}[
colback=gray!3!white,
colframe=gray,
colbacktitle=gray,   
coltitle=white,               
fonttitle=\bfseries,          
title=WorldGenBench-Humanities Prompt,
left=1.5mm,
right=1.5mm,
top=1mm,
bottom=1mm]
\small\ttfamily
In February 1891, at the height of the Chilean Civil War, an aging saltpeter
miner stood in the port of Antofagasta and watched the last shipment of
saltpeter being loaded. As a businessman who supported President Balmaceda, he
was well aware that the Congressional Army was about to occupy this important
port. In the early morning chill, he watched harbor workers come and go, and in
the distance came the sound of a train whistle.
\end{tcolorbox}

The prompt requires port-infrastructure details and regional historical
grounding: Victorian iron cranes, labor clothing, steam locomotives and tracks,
saltpeter packed as white crystals in burlap bags, British-influenced formal
attire for the mine owner, colonial masonry harbor buildings, steam freighters,
the Atacama desert landscape, morning coastal fog, and British merchant-bank
signage. Our RL checkpoint obtains KCS 0.500, while Unify-Agent obtains 0.300
and GenSearcher obtains 0.200. Our result satisfies five checklist points.
First, the port workers wear rough shirts, loose trousers, and sun hats.
Second, the harbor buildings use a colonial masonry style. Third, a steam
freighter is visible in the distance. Fourth, the surrounding terrain is
sparse and desert-like, matching Antofagasta's Atacama setting. Finally, the
sky appears misty in the early morning. The remaining missed points are
Victorian iron cranes, visible steam locomotives and tracks, white crystalline
saltpeter in burlap bags, the mine owner's British formal attire and pocket
watch, and a British merchant-bank sign on the pier.

\begin{table}[!htbp]
\centering
\caption{\textbf{Checklist evaluation for the Antofagasta case.} Each row is
one score point from the benchmark checklist. Blue rows mark the points
satisfied by our model. \ding{51} means satisfied and \ding{55} means not
satisfied.}
\label{tab:worldgen_antofagasta_checklist}
\scriptsize
\setlength{\tabcolsep}{5pt}
\renewcommand{\arraystretch}{1.12}
\resizebox{\linewidth}{!}{%
\begin{tabular}{lccc}
\toprule
Checklist item & Ours & Unify-Agent & GenSearcher \\
\midrule
Harbor docks show typical Victorian iron cranes & \ding{55} & \ding{55} & \ding{55} \\
\rowcolor[HTML]{e8f0fe}
Workers wear rough shirts, baggy pants, and sun hats & \ding{51} & \ding{55} & \ding{55} \\
Steam locomotives and railroad tracks are visible in the scene & \ding{55} & \ding{55} & \ding{55} \\
Saltpeter shipments appear as white crystals packed in burlap bags & \ding{55} & \ding{55} & \ding{55} \\
The mine owner wears British formal attire and a pocket watch & \ding{55} & \ding{55} & \ding{55} \\
\rowcolor[HTML]{e8f0fe}
Harbor buildings are colonial-style masonry structures & \ding{51} & \ding{51} & \ding{55} \\
\rowcolor[HTML]{e8f0fe}
A steam freighter at anchor can be seen in the distance & \ding{51} & \ding{51} & \ding{51} \\
\rowcolor[HTML]{e8f0fe}
The port surroundings show a typical desert landscape & \ding{51} & \ding{51} & \ding{55} \\
\rowcolor[HTML]{e8f0fe}
The sky appears misty in the early morning & \ding{51} & \ding{55} & \ding{51} \\
A British merchant-bank sign is visible on the pier & \ding{55} & \ding{55} & \ding{55} \\
\midrule
KCS & 0.500 & 0.300 & 0.200 \\
\bottomrule
\end{tabular}
}
\end{table}
\FloatBarrier

\begin{figure}[!htbp]
\centering
\includegraphics[width=\linewidth]{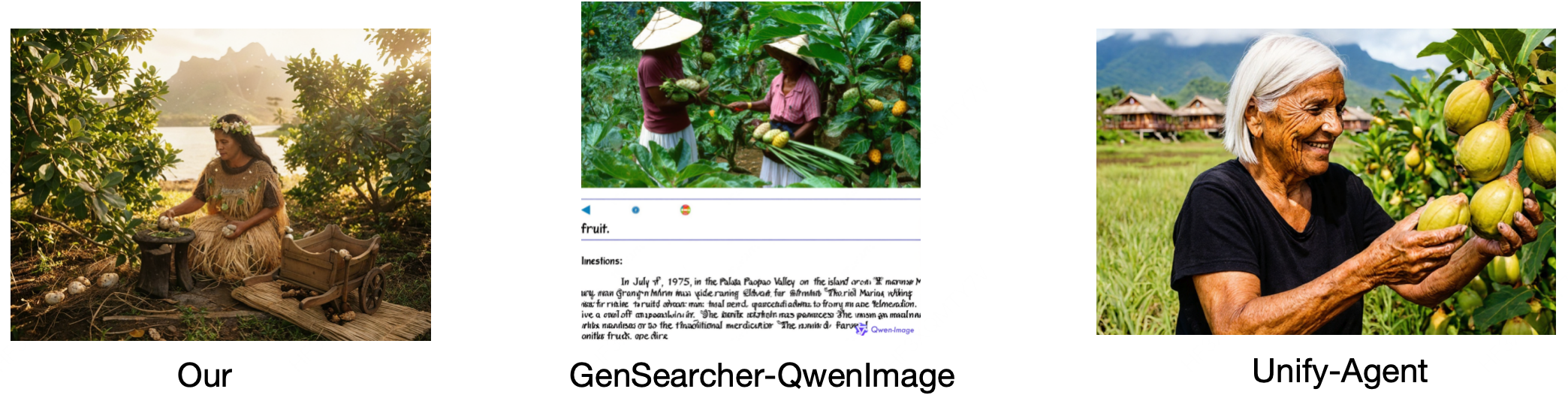}
\caption{Case study on WorldGenBench-Humanities. Our model better grounds the
Moorea noni-harvesting scene in volcanic valley terrain, scattered ripe noni
fruit, a local flower garland, and a distant Pacific view.}
\label{fig:worldgen_moorea_case}
\end{figure}

\cref{fig:worldgen_moorea_case} presents a WorldGenBench-Humanities case about
elderly Granny Maria picking noni fruit in the Paopao Valley on Moorea in July
1975. The full evaluation prompt is:

\begin{tcolorbox}[
colback=gray!3!white,
colframe=gray,
colbacktitle=gray,   
coltitle=white,               
fonttitle=\bfseries,          
title=WorldGenBench-Humanities Prompt,
left=1.5mm,
right=1.5mm,
top=1mm,
bottom=1mm]
\small\ttfamily
In July 1975, in the Paopao Valley on the island of Moorea, elderly Granny
Maria was picking noni fruit. The ripe noni fruits from this orchard, which has
been passed down from generation to generation, give off a special odor. She is
going to make the fruits into a traditional medicine for the upcoming Harvest
Festival celebrations.
\end{tcolorbox}

The prompt requires both agricultural activity and Polynesian island grounding:
volcanic valley terrain, traditional floral clothing, ripe white noni fruit,
a woven rattan basket, breadfruit and banana trees, a local flower garland,
stone orchard fences, red laterite soil, traditional storage sheds, and a
distant Pacific view. Our RL checkpoint obtains KCS 0.400, while Unify-Agent
obtains 0.100 and GenSearcher obtains 0.000. Our result satisfies four
checklist points. First, the background shows steep volcanic island valley
terrain. Second, white ripe noni fruit is scattered on the ground. Third, the
character wears a local flower garland. Finally, the distant view includes the
Pacific Ocean, grounding the scene on Moorea. The remaining missed points are
traditional floral dress, a woven rattan basket, breadfruit and banana trees,
stone fences, red laterite soil, and traditional storage sheds.

\begin{table}[!htbp]
\centering
\caption{\textbf{Checklist evaluation for the Moorea case.} Each row is one
score point from the benchmark checklist. Blue rows mark the points satisfied
by our model. \ding{51} means satisfied and \ding{55} means not satisfied.}
\label{tab:worldgen_moorea_checklist}
\scriptsize
\setlength{\tabcolsep}{5pt}
\renewcommand{\arraystretch}{1.12}
\resizebox{\linewidth}{!}{%
\begin{tabular}{lccc}
\toprule
Checklist item & Ours & Unify-Agent & GenSearcher \\
\midrule
\rowcolor[HTML]{e8f0fe}
Scenery reflects typical volcanic island valley terrain & \ding{51} & \ding{51} & \ding{55} \\
Characters wear traditional Polynesian flower dresses & \ding{55} & \ding{55} & \ding{55} \\
\rowcolor[HTML]{e8f0fe}
The ground is strewn with white ripe noni fruit & \ding{51} & \ding{55} & \ding{55} \\
The character carries a woven rattan basket & \ding{55} & \ding{55} & \ding{55} \\
Breadfruit and banana trees are visible in the scene & \ding{55} & \ding{55} & \ding{55} \\
\rowcolor[HTML]{e8f0fe}
The character wears a local flower garland & \ding{51} & \ding{55} & \ding{55} \\
A stone fence divides the orchard & \ding{55} & \ding{55} & \ding{55} \\
The ground shows laterite formed by weathered volcanic rocks & \ding{55} & \ding{55} & \ding{55} \\
Traditional storage sheds are visible in the scene & \ding{55} & \ding{55} & \ding{55} \\
\rowcolor[HTML]{e8f0fe}
The Pacific Ocean is visible in the distance & \ding{51} & \ding{55} & \ding{55} \\
\midrule
KCS & 0.400 & 0.100 & 0.000 \\
\bottomrule
\end{tabular}
}
\end{table}
\FloatBarrier

\begin{figure}[!htbp]
\centering
\includegraphics[width=\linewidth]{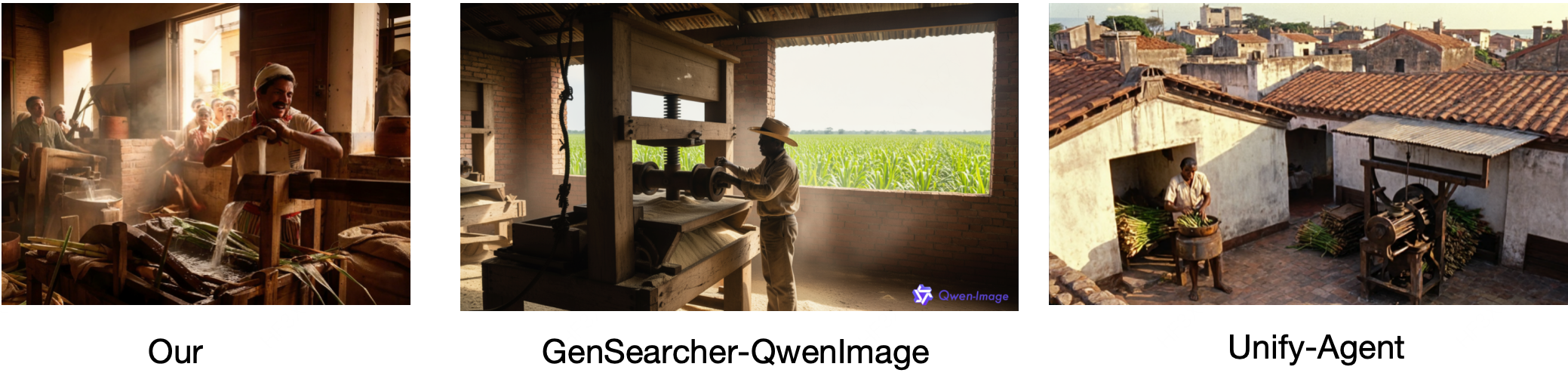}
\caption{Case study on WorldGenBench-Humanities. Our model better grounds the
Santo Domingo sugar-workshop scene in the wooden cane press, earthenware vats,
visible steam from the sugar cooker, and Caribbean-style exterior buildings.}
\label{fig:worldgen_santo_domingo_case}
\end{figure}

\cref{fig:worldgen_santo_domingo_case} presents a WorldGenBench-Humanities case
about a sugar worker in an ancestral workshop in Santo Domingo in 1961. The
full evaluation prompt is:

\begin{tcolorbox}[
colback=gray!3!white,
colframe=gray,
colbacktitle=gray,   
coltitle=white,               
fonttitle=\bfseries,          
title=WorldGenBench-Humanities Prompt,
left=1.5mm,
right=1.5mm,
top=1mm,
bottom=1mm]
\small\ttfamily
On a hot afternoon in 1961, Luis, a sugar worker in the old town of Santo
Domingo, is busy in his ancestral sugar workshop. The air is filled with the
sweet smell of fresh sugar cane from the Caripo Valley. Suddenly, there is a
noise in the distance, as news of the recent fall of the Trujillo regime
reaches the old neighborhood.
\end{tcolorbox}

The prompt requires both traditional sugar-production details and Santo
Domingo's historical setting: a wooden sugar-cane press, sweat-stained white
cotton work clothing, Spanish colonial limestone walls, bagasse on the floor,
earthenware vats, arched doors and windows, Trujillo-period leader portraits,
steam from the sugar cooker, Caribbean-style buildings outside the window, and
bamboo sieves and barrels. Our RL checkpoint obtains KCS 0.400, while
Unify-Agent obtains 0.200 and GenSearcher obtains 0.100. Our result satisfies
four checklist points. First, the workshop includes a traditional wooden
sugar-cane press. Second, earthenware vats used for collecting cane sugar are
visible. Third, steam rises from the sugar cooker, making the production
process concrete. Finally, Caribbean-style buildings appear outside the
window, grounding the scene in Santo Domingo. The remaining missed points are
the sweat-stained white cotton shirt, Spanish colonial limestone walls,
bagasse on the floor, arched doors and windows, Trujillo-period leader
portraits, and bamboo sieves and barrels.

\begin{table}[!htbp]
\centering
\caption{\textbf{Checklist evaluation for the Santo Domingo case.} Each row is
one score point from the benchmark checklist. Blue rows mark the points
satisfied by our model. \ding{51} means satisfied and \ding{55} means not
satisfied.}
\label{tab:worldgen_santo_domingo_checklist}
\scriptsize
\setlength{\tabcolsep}{5pt}
\renewcommand{\arraystretch}{1.12}
\resizebox{\linewidth}{!}{%
\begin{tabular}{lccc}
\toprule
Checklist item & Ours & Unify-Agent & GenSearcher \\
\midrule
\rowcolor[HTML]{e8f0fe}
The workshop has a traditional wooden sugar-cane press & \ding{51} & \ding{51} & \ding{51} \\
The worker wears a sweat-stained white cotton short-sleeved shirt & \ding{55} & \ding{55} & \ding{55} \\
The walls use typical Spanish colonial limestone construction & \ding{55} & \ding{55} & \ding{55} \\
Bagasse is scattered on the ground & \ding{55} & \ding{55} & \ding{55} \\
\rowcolor[HTML]{e8f0fe}
Earthenware vats used to collect cane sugar are visible & \ding{51} & \ding{55} & \ding{55} \\
Windows and doors use a typical Spanish arched design & \ding{55} & \ding{55} & \ding{55} \\
Trujillo-period leader portraits are visible on the walls & \ding{55} & \ding{55} & \ding{55} \\
\rowcolor[HTML]{e8f0fe}
Steam can be seen rising from the sugar cooker & \ding{51} & \ding{55} & \ding{55} \\
\rowcolor[HTML]{e8f0fe}
Typical Caribbean-style buildings can be seen outside the window & \ding{51} & \ding{51} & \ding{55} \\
Bamboo sieves and barrels are placed in the workshop & \ding{55} & \ding{55} & \ding{55} \\
\midrule
KCS & 0.400 & 0.200 & 0.100 \\
\bottomrule
\end{tabular}
}
\end{table}
\FloatBarrier

\subsection{Additional Trajectory Cases}
\label{app:additional_trajectories}
The following case follows the same display format as the main-text case
study. The constant system prompt is omitted. For each remaining turn, the
content is extracted from the original ourtraj: reasoning text is shown
without \texttt{<think>} tags, tool calls are shown as tool names and query
lists rather than raw JSON/XML wrappers, tool observations are retained, and
the final generation caption and image are shown at the end.

\begin{tcolorbox}[
    promptbox,
    title={Case Study: Whitechapel, London},
    breakable,
    fontupper=\small\ttfamily,
    boxsep=2pt,
    top=3pt,
    bottom=3pt,
    left=5pt,
    right=5pt
]
\setlength{\parindent}{0pt}
\setlength{\parskip}{1pt}

\begin{center}
    \includegraphics[width=\linewidth,keepaspectratio]{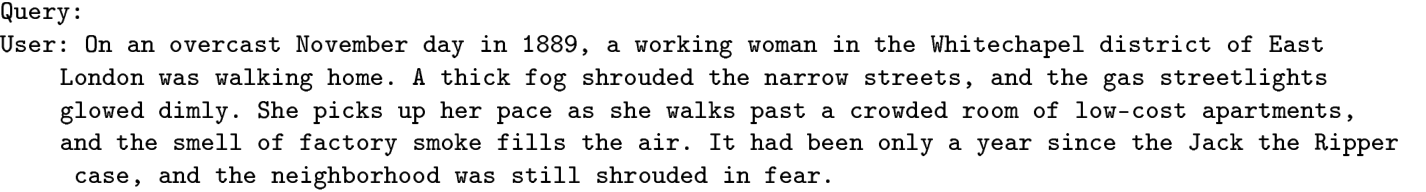}
\end{center}

\begin{center}
    \includegraphics[width=\linewidth,keepaspectratio]{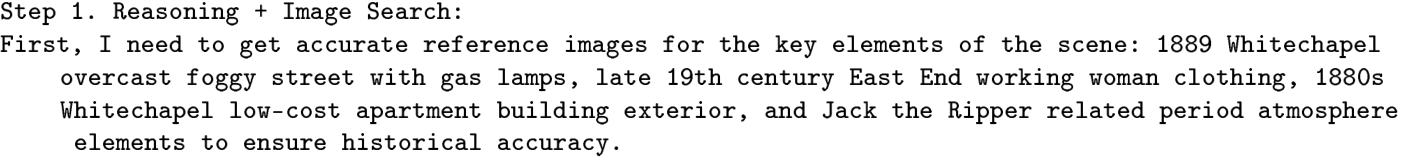}
\end{center}

\begin{center}
    \includegraphics[width=\linewidth,keepaspectratio]{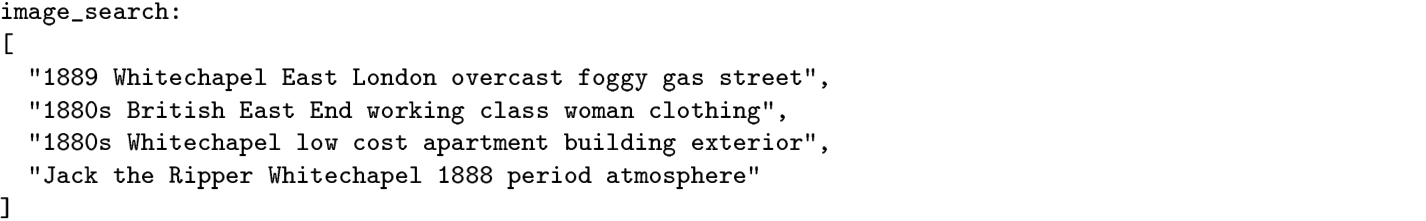}
\end{center}

\begin{center}
    \includegraphics[width=\linewidth,keepaspectratio]{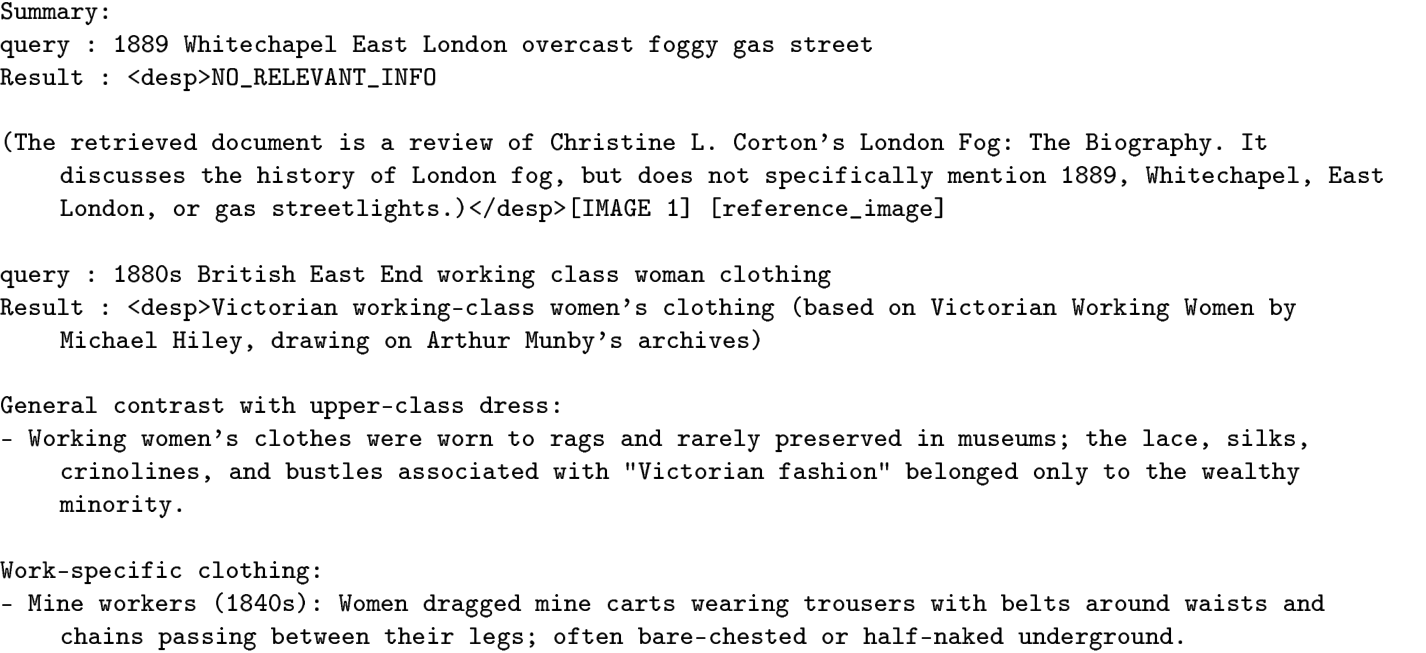}
\end{center}

\begin{center}
    \includegraphics[width=\linewidth,keepaspectratio]{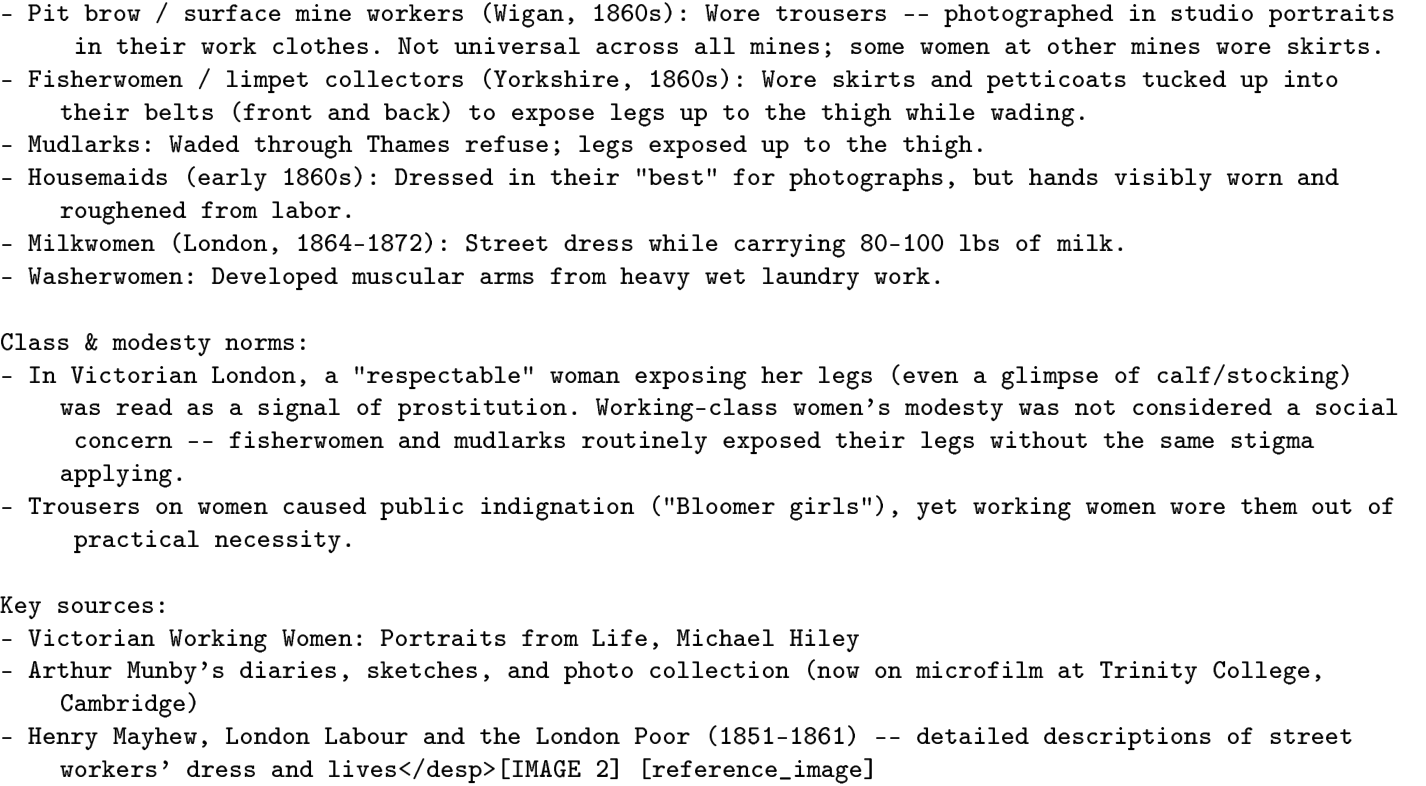}
\end{center}

\begin{center}
    \includegraphics[width=\linewidth,keepaspectratio]{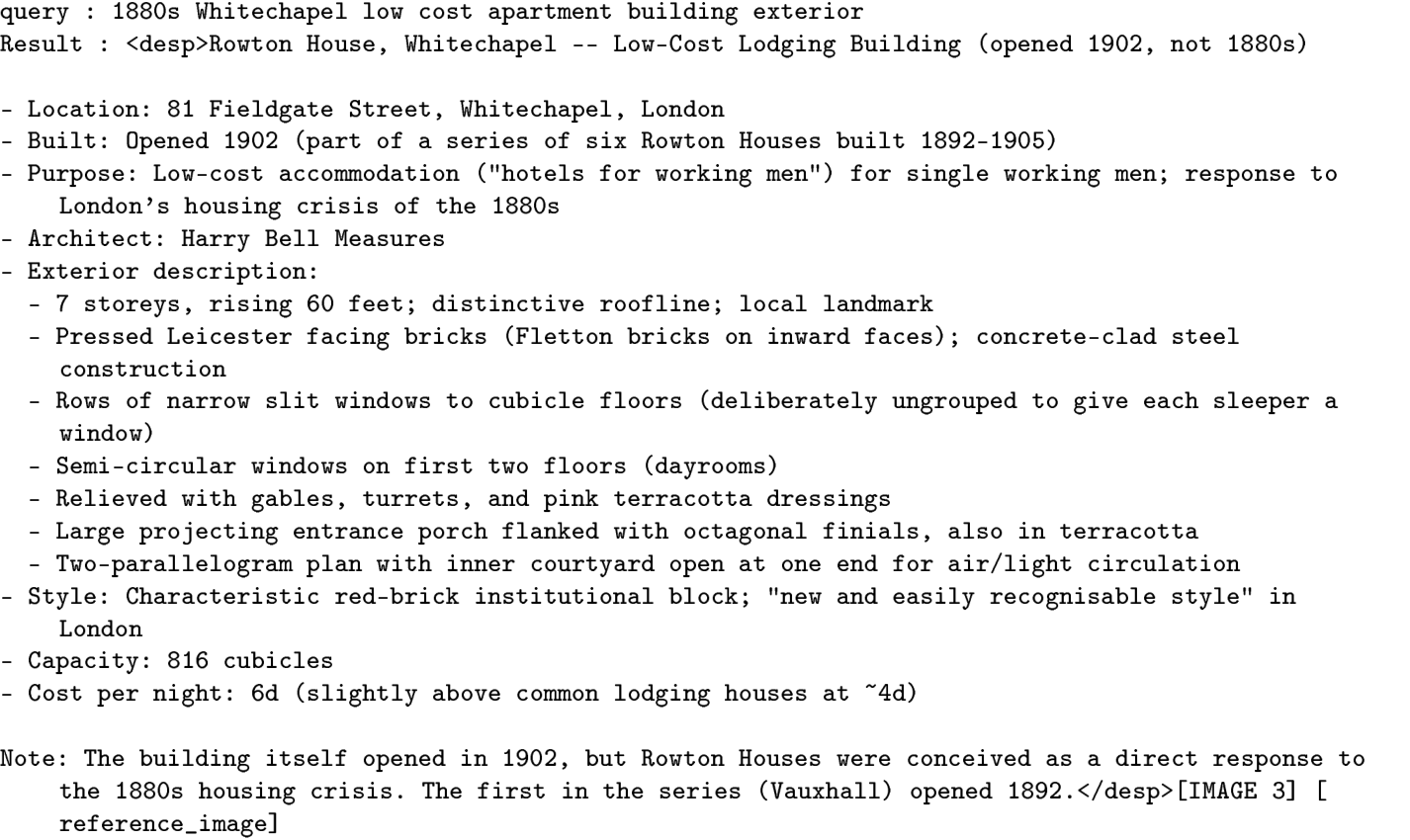}
\end{center}

\begin{center}
    \includegraphics[width=\linewidth,keepaspectratio]{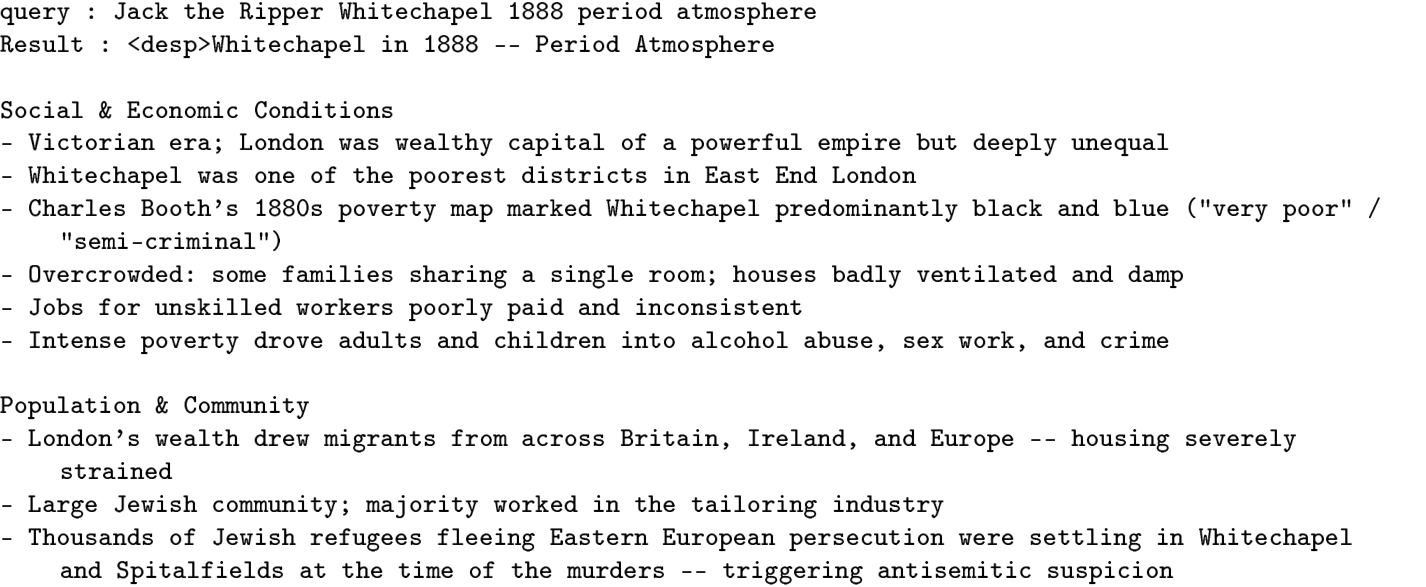}
\end{center}

\begin{center}
    \includegraphics[width=\linewidth,keepaspectratio]{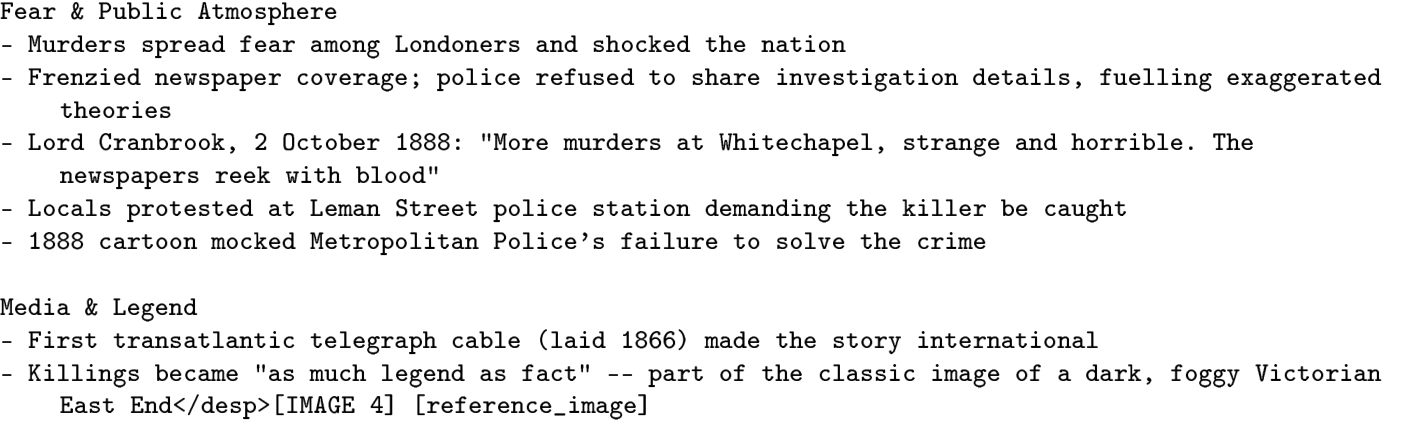}
\end{center}

\begin{center}
    \includegraphics[width=\linewidth,keepaspectratio]{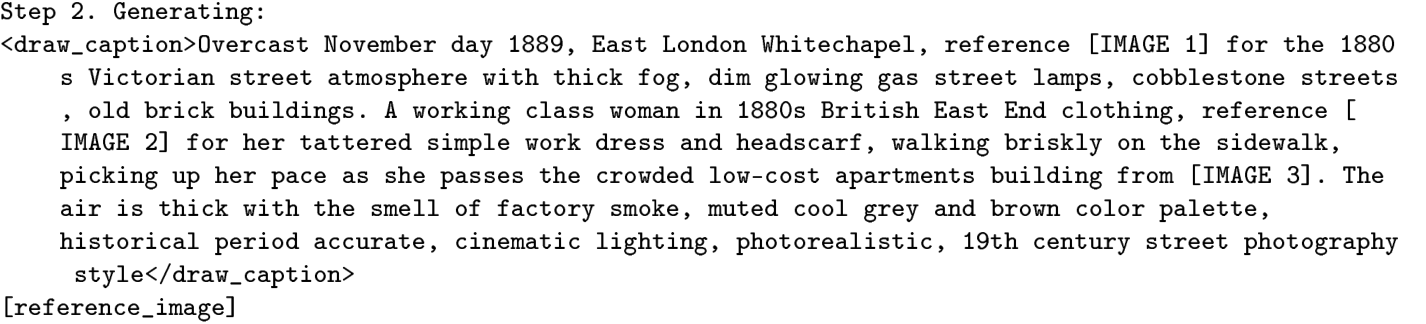}
\end{center}

\begin{center}
    \includegraphics[width=0.4\linewidth,keepaspectratio]{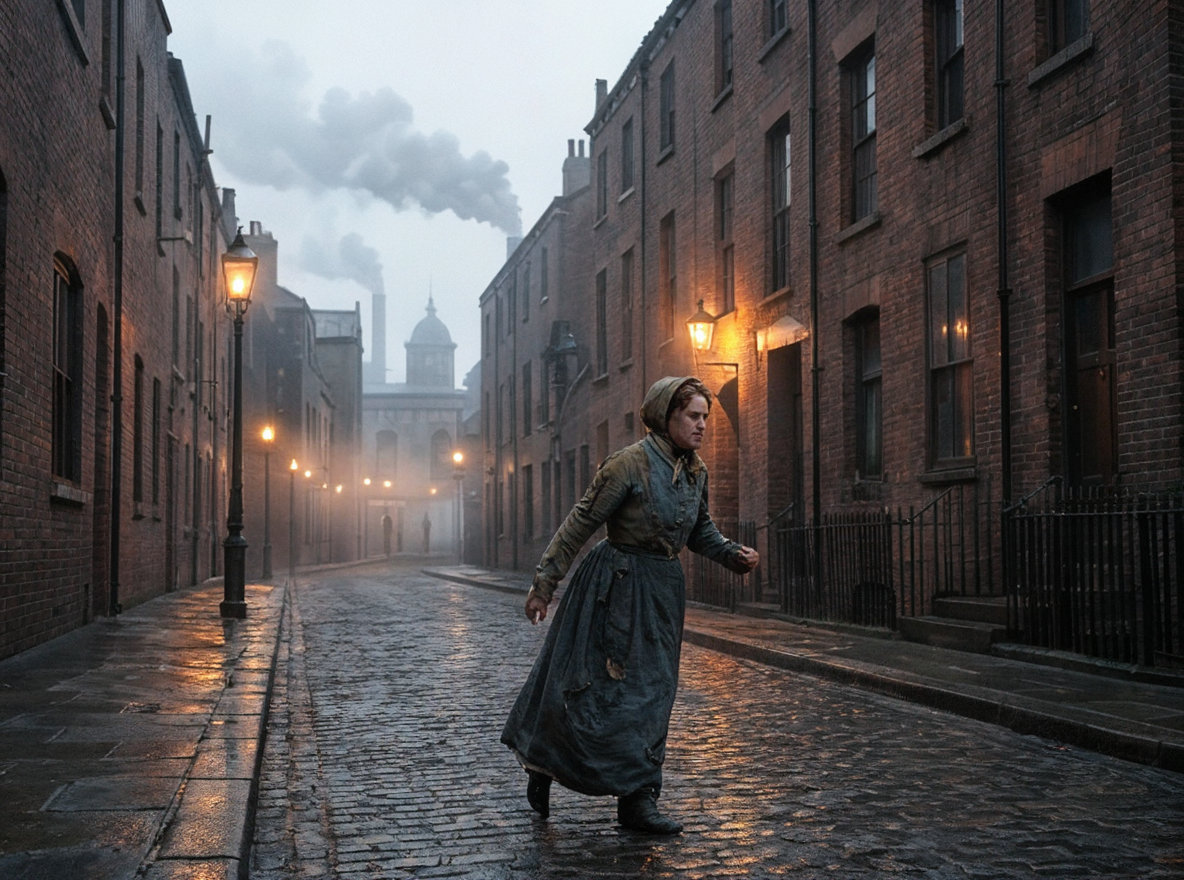}
\end{center}

\end{tcolorbox}

\subsection{SFT Data Examples}
\label{app:sft_examples}
The following SFT rollout starts at the user question. The system prompt is omitted.
The original text is kept, while image placeholders are replaced by the actual
reference and generated images.

\begin{tcolorbox}[
    promptbox,
    title={SFT Example: WGS 84 vs GRS 80 Raw Rollout},
    breakable,
    fontupper=\small\ttfamily,
    boxsep=2pt,
    top=3pt,
    bottom=3pt,
    left=5pt,
    right=5pt
]
\setlength{\parindent}{0pt}
\setlength{\parskip}{2pt}

[message 1 | role=user]

User: A technical diagram comparing the reference ellipsoids of 'WGS 84' and
'GRS 80'; the illustration highlights the difference in the semi-minor axis. A
data plaque lists the exact length of the semi-minor axis in meters for WGS 84
and the name of the international organization that maintains the standard; both
the length and the organization name must be correct.

[message 2 | role=assistant]

\textless{}think\textgreater{}
First, I need to obtain the accurate exact semi-minor axis values for WGS 84 and
GRS 80, as well as the official organization that maintains the WGS 84 standard,
to ensure the data on the diagram is fully correct. I will use text search to get
this authoritative information first.
\textless{}/think\textgreater{}

\textless{}tool\_call\textgreater{}
\{
  "name": "text\_search",
  "arguments": \{
    "query": [
      "WGS84 GRS80 reference ellipsoid semi-minor axis exact length meters",
      "WGS84 standard maintaining international organization",
      "GRS80 semi-minor axis precise value meters"
    ]
  \}
\}
\textless{}/tool\_call\textgreater{}

[message 3 | role=tool]

query : WGS84 GRS80 reference ellipsoid semi-minor axis exact length meters

Result : World Geodetic System - Wikipedia | Semi-minor axis (b) comparison -
WGS 84 vs GRS 80:

GRS 80: semi-major axis 6,378,137.0 m; semi-minor axis ~ 6,356,752.314140 m.
WGS 84: semi-major axis 6,378,137.0 m; semi-minor axis ~ 6,356,752.314245 m.

Both share the same semi-major axis, but differ in inverse flattening, resulting
in a 0.105 mm difference in the semi-minor axis. WGS 84 semi-minor axis is a
computed parameter: b = a x (1 - f) = 6,356,752.3142 m.

query : WGS84 standard maintaining international organization

Result : WGS84 - World Geodetic System 1984 - geodesy.science - IAG website |
WGS84 is maintained by the U.S. Department of Defense, specifically the National
Geospatial-Intelligence Agency (NGA). The document also states that WGS84 is
aligned with ITRF and that ITRF is maintained under the broader framework of
global geodesy, with IAG and GGOS playing key roles.

query : GRS80 semi-minor axis precise value meters

Result : NOAA Technical Memorandum NOS NGS 97 Storage and Use of GRS80 Ellipsoid
Parameters in the Modernized National Spatial Reference System.

Formula (11a): b = a(1 - 1/f\string^-1). Substituting a = 6378137 m and
f\string^-1 = 298.257222101 gives b ~= 6356752.3141 m.

[message 4 | role=assistant]

\textless{}think\textgreater{}
Now I have the accurate data points needed: WGS 84 semi-minor axis =
6356752.3142 m, maintained by the U.S. National Geospatial-Intelligence Agency
(NGA); GRS 80 semi-minor axis = 6356752.3141 m, with a 0.105 mm difference
between the two. Next I need to search for reference technical diagrams comparing
reference ellipsoids, to get the correct visual structure for the illustration.
\textless{}/think\textgreater{}

\textless{}tool\_call\textgreater{}
\{
  "name": "image\_search",
  "arguments": \{
    "query": [
      "WGS84 GRS80 reference ellipsoid comparison technical diagram",
      "reference ellipsoid semi-minor axis illustration technical diagram",
      "geodesy ellipsoid comparison diagram with data plaque"
    ]
  \}
\}
\textless{}/tool\_call\textgreater{}

[message 5 | role=tool]

query : WGS84 GRS80 reference ellipsoid comparison technical diagram

Result : WGS84 vs GRS80 reference ellipsoid comparison data

Both share the same semi-major axis: 6 378 137 m. Semi-minor axis:
GRS 80 = 6 356 752.3141 m; WGS 84 = 6 356 752.3142 m. The flattening differs by
a tiny amount, so the maximum ellipsoid height difference at the pole is only
0.1 mm.

[external image 1]

\begin{center}
\includegraphics[width=0.4\linewidth,keepaspectratio]{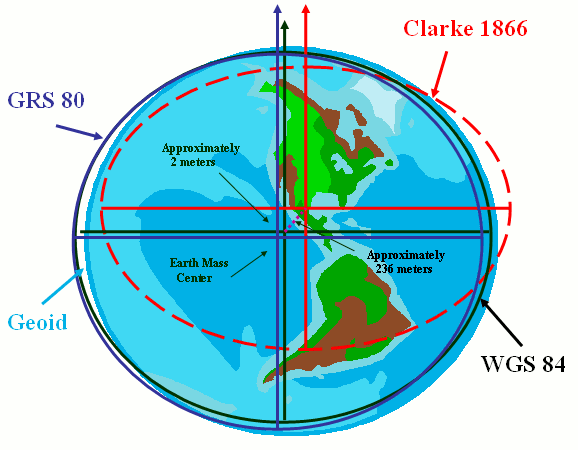}
\end{center}

query : reference ellipsoid semi-minor axis illustration technical diagram

Result : Reference Ellipsoid Semi-minor Axis - Key Technical Information

Semi-minor axis = half of the minor axis of an ellipse; for a reference
ellipsoid, it is the north-south axis around which the ellipse is rotated to
form a solid. The ellipsoid bulges at the equator and is flattened at the poles.
An ellipsoid is completely defined by its semi-major axis and flattening.

[external image 2]

\begin{center}
\includegraphics[width=0.4\linewidth,keepaspectratio]{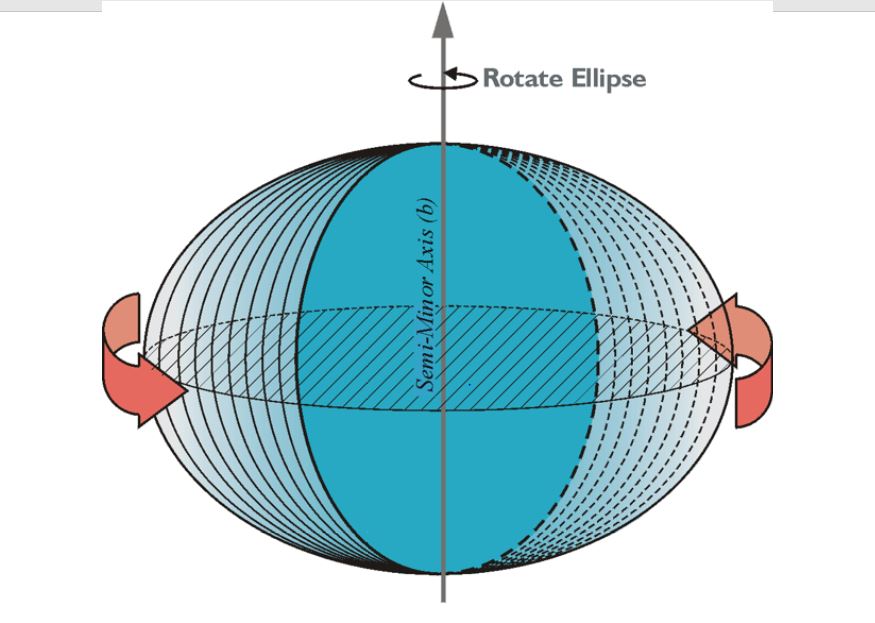}
\end{center}

query : geodesy ellipsoid comparison diagram with data plaque

Result : GRS80 vs WGS84 Ellipsoid Comparison

Both ellipsoids have semi-major axis 6,378,137 m. GRS80 inverse flattening is
298.257222101; WGS84 inverse flattening is 298.257223563. GRS80 was adopted by
IUGG in 1979 and WGS84 is the reference for the GPS datum.

[external image 3]

\begin{center}
\includegraphics[width=0.4\linewidth,keepaspectratio]{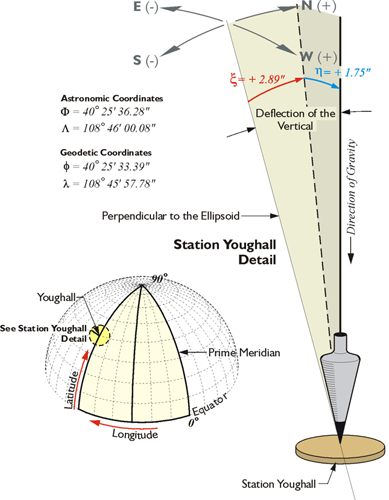}
\end{center}

[message 6 | role=assistant]

\textless{}think\textgreater{}
Okay, now I have the reference diagrams and all the correct data. I need to
create a technical diagram that compares WGS84 and GRS80 reference ellipsoids,
highlights the semi-minor axis difference, and has a data plaque with the correct
numbers and maintenance org.

First, use the first reference image [IMAGE1] as the base for the ellipsoid
comparison layout, the second [IMAGE2] for the semi-minor axis labeling. Then the
prompt should specify: technical diagram, side-by-side comparison (exaggerate the
tiny semi-minor axis difference for visibility, note it's 0.105mm difference),
label semi-minor axis clearly for both, a data plaque with: WGS84 semi-minor axis
= 6356752.3142 m, maintained by US National Geospatial-Intelligence Agency (NGA);
GRS80 semi-minor axis = 6356752.3141 m; note that semi-major axis is identical
6378137.0 m for both, clean white background, engineering/geodesy technical
illustration style, clear labels, no clutter.

Yes that makes sense. Let's call the draw tool now.
\textless{}/think\textgreater{}

\textless{}tool\_call\textgreater{}
\{
  "name": "draw",
  "arguments": \{
    "prompt": "Technical geodesy diagram comparing WGS 84 and GRS 80 reference ellipsoids, based on the layout of [IMAGE1] and using the semi-minor axis labeling style of [IMAGE2]. Exaggerate the tiny 0.105 mm difference in semi-minor axis for visibility (with a small note that the difference is not to scale). Clearly label the semi-minor axis (polar radius) for both ellipsoids with callouts. Include a formal data plaque at the bottom right listing: 1. WGS 84: Semi-minor axis = 6356752.3142 m, maintained by the U.S. National Geospatial-Intelligence Agency (NGA); 2. GRS 80: Semi-minor axis = 6356752.3141 m; 3. Shared semi-major axis = 6378137.0 m. Clean white background, professional technical illustration style, sharp clear text labels, no extraneous elements, all numerical values accurate.",
    "images": [
      "https://vdatum.noaa.gov/images/docs/clarke1866\_wgs84\_grs80.gif",
      "https://courses.ems.psu.edu/geog862/sites/geog862/files/images/Lesson05/Ellipsoid.JPG"
    ]
  \}
\}
\textless{}/tool\_call\textgreater{}

[message 7 | role=tool]

Result : [internal image 4]

\begin{center}
\includegraphics[width=0.4\linewidth,keepaspectratio]{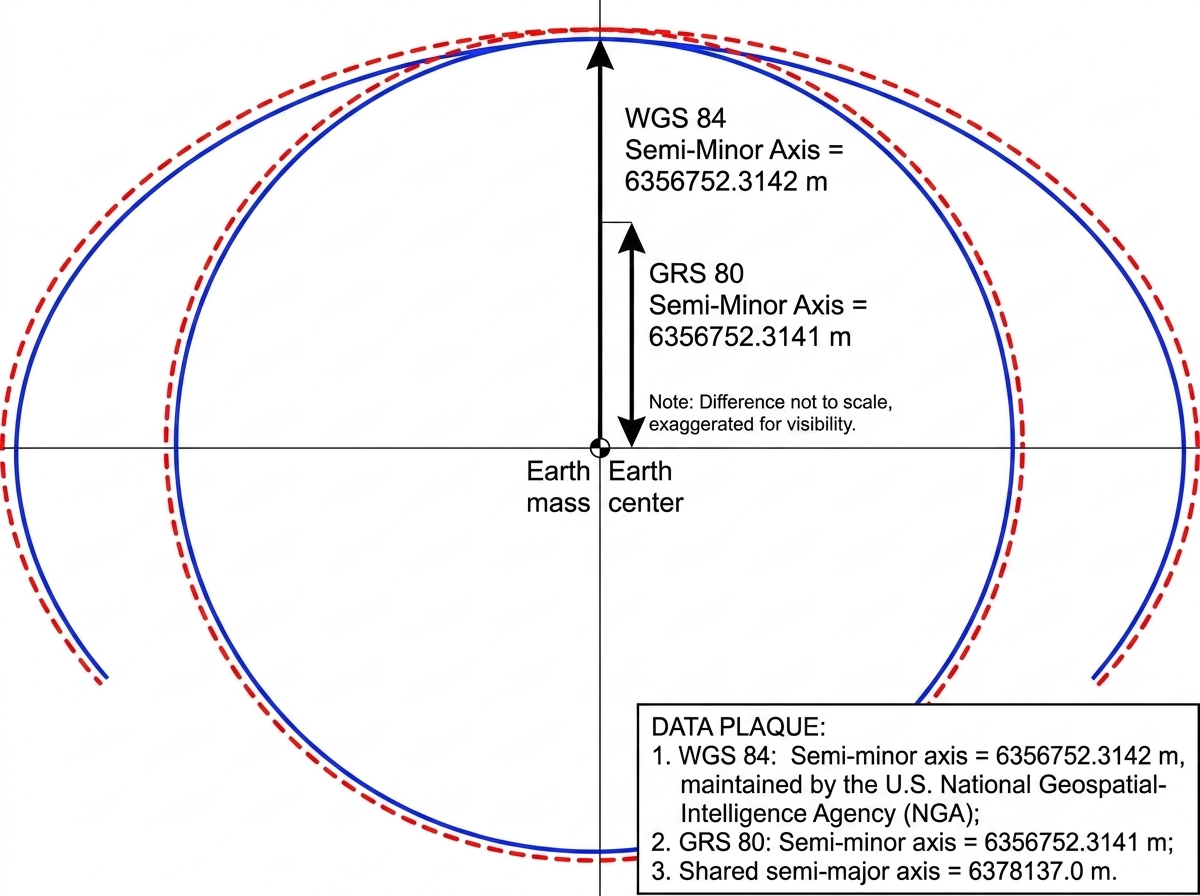}
\end{center}

[message 8 | role=assistant]

\textless{}think\textgreater{}
Got it, let's check the generated image first. First, the semi-minor axis values
are correct: WGS84 is 6356752.3142 m, GRS80 is 6356752.3141 m, right. The data
plaque has the correct organization: U.S. National Geospatial-Intelligence
Agency (NGA) for WGS84, perfect. The difference is exaggerated with a note that
it's not to scale, which is good because the actual difference is only 0.1mm, you
can't see it otherwise. The shared semi-major axis is listed correctly as
6378137.0 m. The labels are clear, it's a proper technical diagram, the
semi-minor axis is highlighted as requested. All requirements are met. Now I just
need to output the final result with the path, and a short description.
\textless{}/think\textgreater{}

\textless{}\textbar{}box\_start\textbar{}\textgreater{}Generated Technical Diagram: WGS 84 vs GRS 80 Reference Ellipsoid Comparison.
\textless{}\textbar{}box\_end\textbar{}\textgreater{}
\end{tcolorbox}

\begin{tcolorbox}[
    promptbox,
    title={SFT Example: Sigewinne and Bailu Raw Rollout},
    breakable,
    enhanced,
    fontupper=\small\ttfamily,
    boxsep=2pt,
    top=3pt,
    bottom=3pt,
    left=5pt,
    right=5pt,
    before upper={\sloppy},
]
\setlength{\parindent}{0pt}
\setlength{\parskip}{2pt}

[message 1 | role=user]

User: In a whimsical underwater nursery, the Melusine nurse Sigewinne from
Genshin Impact and the young dragon-healer Bailu from Honkai: Star Rail are
mixing a glowing blue potion; a medical chart on the wall must correctly
display Sigewinne's official Vision element.

[message 2 | role=assistant]

\textless{}think\textgreater{}
First, I need to confirm Sigewinne's official Vision element, then search for
reference images of both characters and the cute underwater nursery mood so the
final image stays canonically correct and visually coherent.
\textless{}/think\textgreater{}

\textless{}tool\_call\textgreater{}
\{
  "name": "text\_search",
  "arguments": \{
    "query": [
      "Sigewinne Genshin Impact official Vision element",
      "Sigewinne confirmed Vision type Genshin"
    ]
  \}
\}
\textless{}/tool\_call\textgreater{}

[message 3 | role=tool]

Sigewinne's Vision is Hydro.

[message 4 | role=assistant]

\textless{}think\textgreater{}
Now I need image references for Sigewinne, Bailu, the Hydro symbol, and the
glowing potion. I'll search for those and then build the final caption.
\textless{}/think\textgreater{}

\textless{}tool\_call\textgreater{}
\{
  "name": "image\_search",
  "arguments": \{
    "query": [
      "Sigewinne Genshin Impact official character art",
      "Bailu Honkai: Star Rail official character art",
      "Genshin Impact official Hydro Vision symbol",
      "glowing blue magical potion fantasy"
    ]
  \}
\}
\textless{}/tool\_call\textgreater{}

[message 5 | role=tool]

query : Sigewinne Genshin Impact official character art

Query result:
Sigewinne's official character image references and details:
- Hydro Vision
- Bow user
- Melusine nurse from Fontaine
- Character art reference available.

\begin{center}
\includegraphics[width=0.16\linewidth,keepaspectratio]
{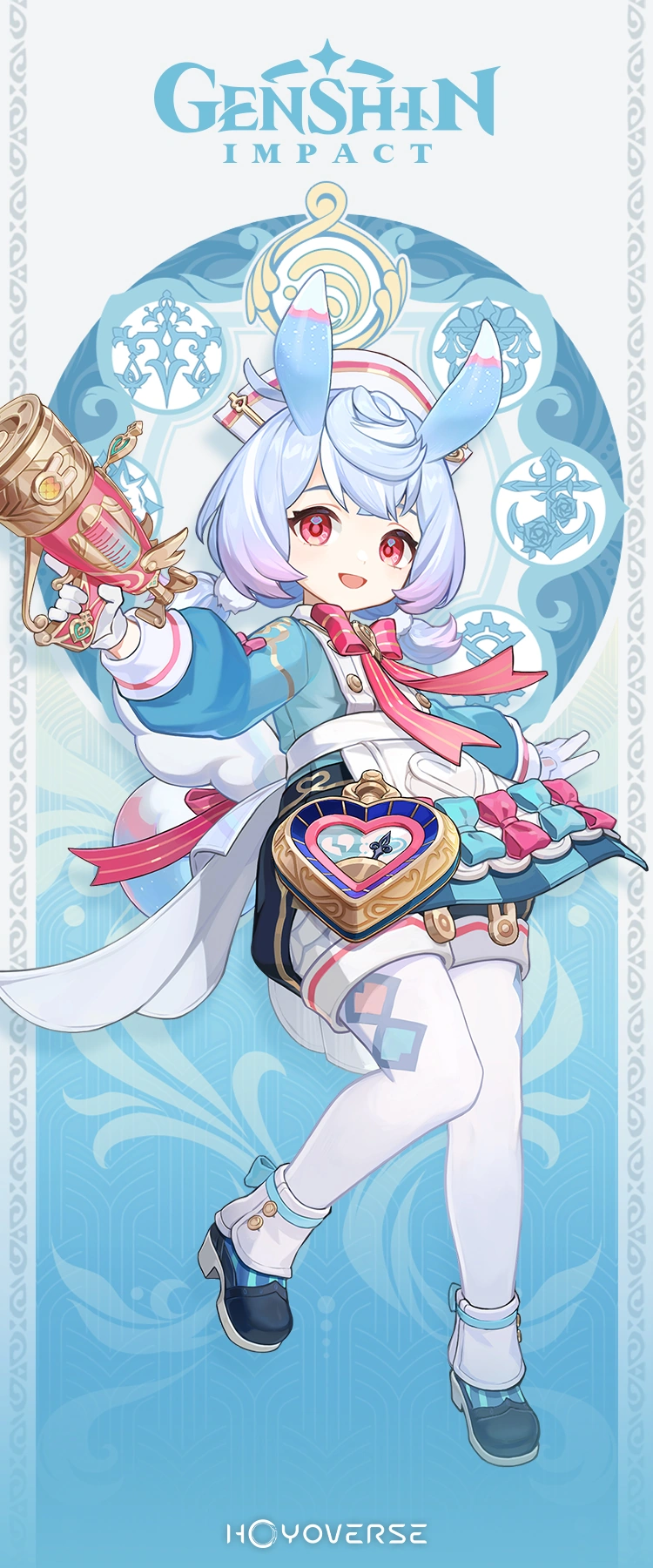}
\end{center}

query : Bailu Honkai: Star Rail official character art

Query result:
Bailu character reference image and appearance details.

\begin{center}
\includegraphics[width=0.35\linewidth,keepaspectratio]
{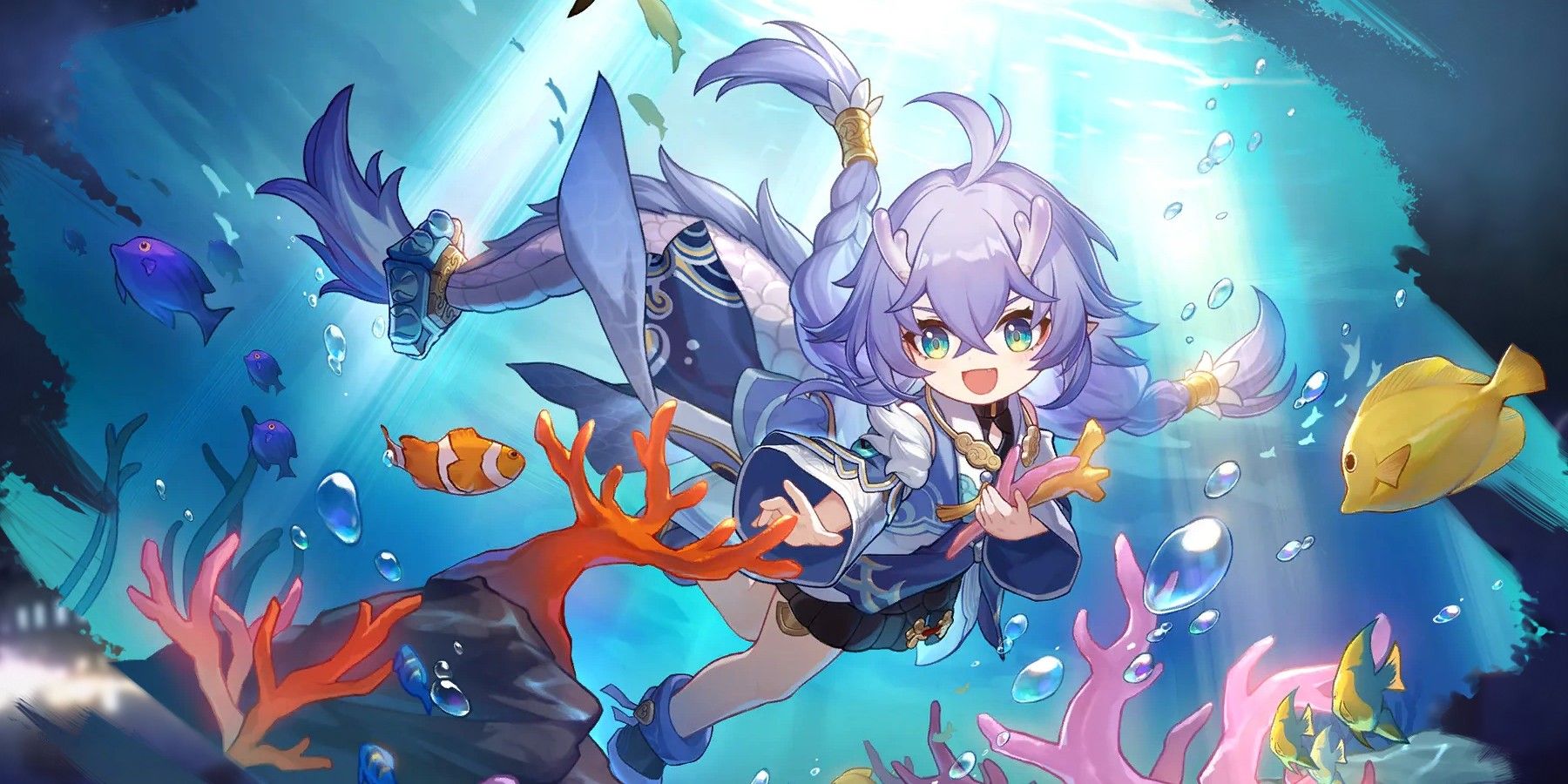}
\end{center}

query : whimsical cute underwater nursery interior design

Query result:
Underwater nursery references with pastel colors, ocean decorations,
and magical cozy atmosphere.

\begin{center}
\includegraphics[width=0.35\linewidth,keepaspectratio]
{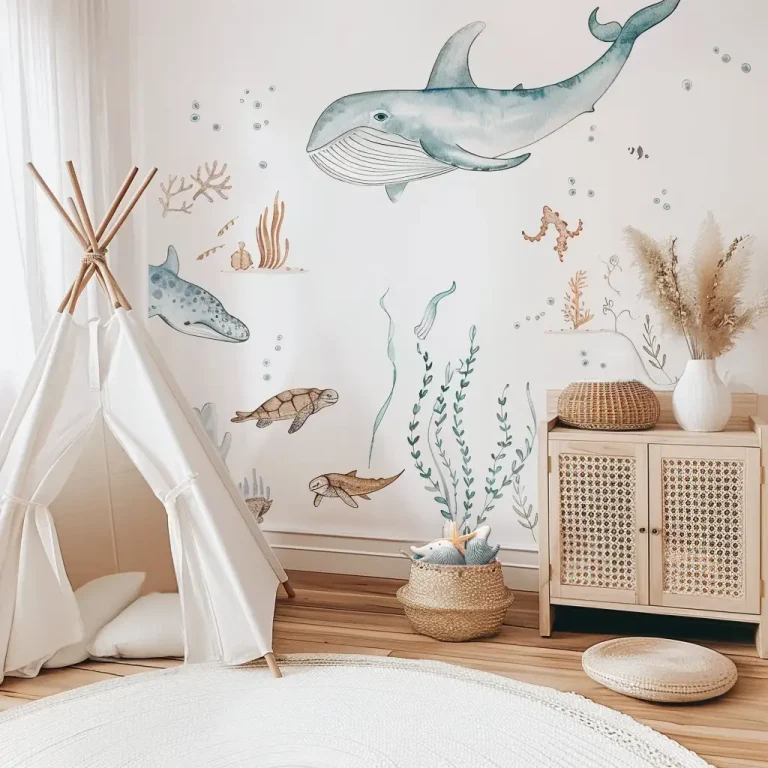}
\end{center}

query : Genshin Impact official Hydro Vision symbol

Query result:
Hydro Vision reference image.

\begin{center}
\includegraphics[width=0.35\linewidth,keepaspectratio]
{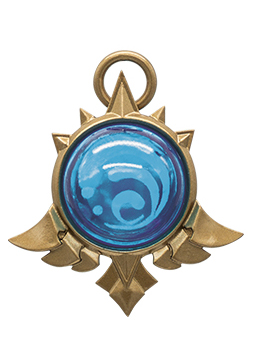}
\end{center}

query : glowing blue magical potion fantasy

Query result:
Blue glowing fantasy potion reference.

\begin{center}
\includegraphics[width=0.35\linewidth,keepaspectratio]
{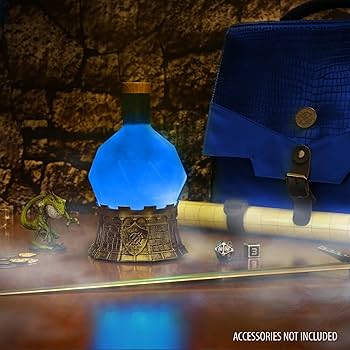}
\end{center}

[message 6 | role=assistant]

\textless{}think\textgreater{}
I have the key references now. I can write a single generation prompt that
uses Sigewinne and Bailu's exact appearances, the Hydro chart, and the glowing
blue potion inside a cute underwater nursery.
\textless{}/think\textgreater{}

\textless{}tool\_call\textgreater{}
\{
  "name": "draw",
  "arguments": \{
    "prompt":
      "Whimsical cozy underwater nursery scene. "
      "Soft blue light filters through the water with small bubbles floating "
      "gently. Walls have cute watercolor sea creature decals, warm light "
      "wood furniture, woven storage baskets, and a pastel color palette."
      "Two cute anime girls stand side by side at a small wooden table: "
      "Sigewinne the Melusine nurse matching [IMAGE 1], and Bailu the young "
      "dragon healer matching [IMAGE 2]. They are mixing a glowing blue "
      "magic potion matching [IMAGE 4]."
      "A framed medical chart is pinned on the wall behind them. The chart "
      "prominently displays the official Hydro Vision symbol matching "
      "[IMAGE 3], correctly indicating Sigewinne's Hydro element."
      "HoYoverse anime aesthetic, soft shading, high detail, 4K resolution, "
      "whimsical magical atmosphere.",
    "images": [
      "sigewinne-ref.png",
      "bailu.jpg",
      "hydro.jpg",
      "potion.jpg"
    ]
  \}
\}
\textless{}/tool\_call\textgreater{}

\begin{center}
\includegraphics[width=0.35\linewidth,keepaspectratio]
{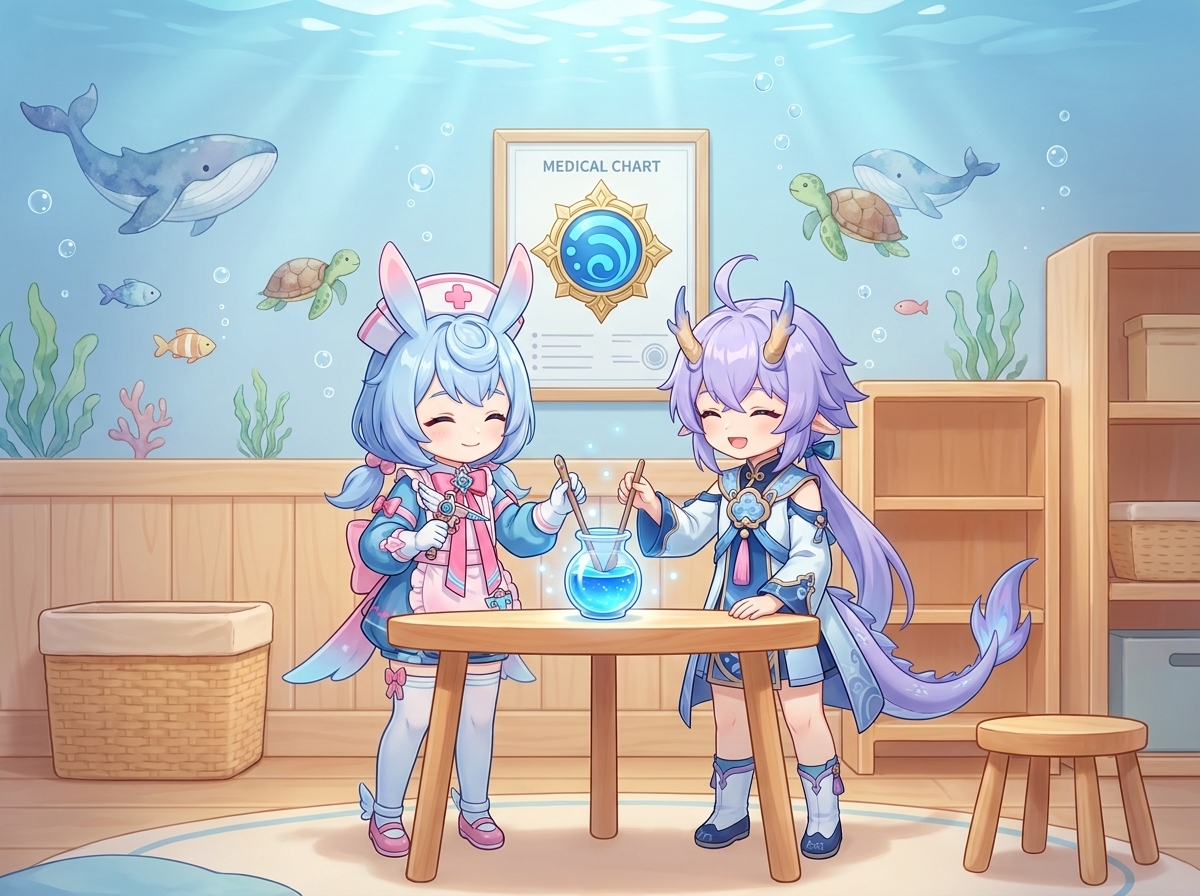}
\end{center}

[message 8 | role=assistant]

\textless{}think\textgreater{}
The result is consistent with the requested anime style. Sigewinne and Bailu
are correctly depicted, and the medical chart shows Hydro as required.
\textless{}/think\textgreater{}

Generated image meets all requirements.

\end{tcolorbox}

\end{document}